\documentclass[runningheads]{llncs}
\usepackage{graphicx}
\usepackage[utf8]{inputenc} 
\usepackage[T1]{fontenc}    
\usepackage{hyperref}       
\usepackage{url}            
\usepackage{booktabs}       
\usepackage{amsfonts}       
\usepackage{nicefrac}       
\usepackage{microtype}      
\usepackage[misc]{ifsym}

\usepackage{amsmath,amsfonts,amsthm,bbm,amssymb}
\usepackage[retain-unity-mantissa = false]{siunitx}
\usepackage{tikz}
\usepackage{pgfplots}
\usepackage{aircraftshapes}
\usetikzlibrary{angles,quotes,shapes.geometric,arrows,matrix}
\usepackage{caption}
\usepackage{subcaption}
\usepackage{enumitem}
\DeclareMathOperator{\E}{\mathbb{E}}

\DeclareMathOperator*{\argmin}{arg\,min}
\DeclareMathOperator*{\argmax}{arg\,max}

\usepackage{algorithm}
\usepackage[noend]{algorithmic}

\makeatletter
\newcommand{\printfnsymbol}[1]{%
  \textsuperscript{\@fnsymbol{#1}}%
}
\makeatother

\begin{document}
\title{Robust Super-Level Set Estimation using Gaussian Processes}
%
%
\author{Andrea Zanette \and
Junzi Zhang \and
Mykel J. Kochenderfer\Letter}
\authorrunning{Andrea Zanette, Junzi Zhang and Mykel J. Kochenderfer}
\tocauthor{Andrea~Zanette, Junzi~Zhang, Mykel~J.~Kochenderfer}
\toctitle{Robust Super-Level Set Estimation using Gaussian Processes}
%
\institute{Stanford University, Stanford, CA, USA\\
\email{\{zanette,junziz,mykel\}@stanford.edu}}
\maketitle              
\begin{abstract}
This paper focuses on the problem of determining as large a region as possible where a function exceeds a given threshold with high probability. We assume that we only have access to a noise-corrupted version of the function and that function evaluations are costly. To select the next query point, we propose maximizing the expected volume of the domain identified as above the threshold as predicted by a Gaussian process, robustified by a variance term. We also give asymptotic guarantees on the exploration effect of the algorithm, regardless of the prior misspecification. We show by various numerical examples that our approach also outperforms existing techniques in the literature in practice.

\end{abstract}
\section{Introduction}
\label{sec:intro}

Many scientific and engineering problems involve determining the maximum value a function $f$  over a region $\Omega$. However, some applications require determining a large subregion of $\Omega$ where the function under consideration exceeds a given threshold $t$. This problem of super-level set estimation arises naturally in the context of safety control, signal coverage, and environmental monitoring \cite{Got13}. 

Formally, we consider the problem of finding the region where a function is above some threshold $t$ with probability at least $\delta$:
\begin{equation}
\{\mathbf x \in \Omega \mid P \left( f( \mathbf x)>t \right) > \delta \}.
\label{introobj}
\end{equation}
We assume that we only have access to a noise-corrupted version of the function and that function evaluations are costly.
In order to fully specify the set in Eqn. (\ref{introobj}), a probabilistic model $P$ for the function must be assumed. In this work, we use Gaussian processes \cite{GPML}, a standard model that directly provides confidence intervals and can easily incorporate new information from the samples. 

The problem is closely related to Bayesian optimization, but the associated techniques are not directly transferable to the problem of level set estimation. A major issue is that it is unclear whether one should focus on identifying points around the threshold for better separation, or aim at points far from the threshold to accelerate the discovery of interested regions. Similar to other exploration-exploitation trade-offs in active learning, we give affirmative answers to both by proposing Robust Maximum Improvement for Level-set Estimation (\textsc{RMILE}), an algorithm to maximize the expected volume of the domain where a function exceeds the threshold with high probability, robustified by a exploration-driven variance term. 

Furthermore, we discuss a criterion for establishing convergence of a generic acquisition function on finite grids that is robust to misspecification of the models. In particular, we show how this criterion applies to our algorithm.  

\paragraph{Related work.}
Some relevant techniques for addressing costly function evaluations are found in the field of Bayesian optimization, which have received growing attention in recent years \cite{Fre16}. Bayesian optimization aims at finding the maximum or minimum of a black box function by repeatedly updating prior beliefs over the function in a Bayesian fashion. This framework is particularly well suited for global optimization of costly functions because it makes effective use of all the information (\textit{i.e.}, the samples) acquired during the process by carefully selecting query points according to an acquisition function. Examples of acquisition functions include the probability of improvement \cite{Kush64}, expected improvement \cite{Mock78}, the Gaussian process upper confidence bound \cite{Sri10}, and information-based policies \cite{shah14}. 

In the literature of level set estimation, when level sets are estimated from an existing dataset, several approaches are available \cite{Wil07},\cite{Son12},\cite{Kri13}. In the context of active learning, a topic of growing interest \cite{Kra08},\cite{Mar17},\cite{Bu09}, one technique is known as the Straddle heuristic \cite{Bry05}, where the expected value and variance given by a Gaussian process are combined to characterize the uncertainty at each candidate point, based on which the next query point is chosen.
 A refinement of the Straddle heuristic was suggested as the \textsc{LSE} algorithm, which is an online classification method based on confidence intervals with some information-theoretic convergence guarantees \cite{Got13}.  This idea is further developed in \cite{Bog16}, with theoretical guarantees that offer a unifying framework of Bayesian optimization and level set estimation.
In a different direction, a Gaussian process-based algorithm addressing time-varying level set estimation was proposed \cite{Yang14}, where a global expected error estimate is adopted as the acquisition function.

A problem similar to ours was considered by \cite{Ma14}, where the authors partition the space using a predefined coarse grid and the goal is to find sub-regions with an average ``score'' above some threshold. The computation of the scores relies on Bayesian quadrature, and no theoretical guarantee on the algorithm is provided. It was later extended to find regions matching general patterns not restricted to the excess of some threshold \cite{Ma15}. Although some preliminary analysis about exploration-exploitation trade-off is given, there is no discussion about the limit behavior of the algorithm.  

In some applications, one is directly concerned with the online estimation of the volume of the super-level set with some given threshold \cite{Bect12},\cite{Cheval14}. However, existing methods are still focused on uncertainty reduction mechanisms similar to the Straddle heuristic. 

This paper is structured as follows. Section 2 provides the necessary preliminaries for our work. Section 3 discusses the relation between super-level set estimation and binary classification, and proposes the \textsc{RMILE} algorithm. The asymptotic behavior of \textsc{RMILE} is discussed in Section 4, followed by various numerical experiments in Section 5 and some conclusive remarks in Section 6.

\section{Preliminaries}
We assume that we only have noisy measurements of the function $f(\mathbf x)$. In other words, we have access to $f(\mathbf x) + \epsilon$ where the noise $\epsilon$ is normally distributed $\epsilon \sim \mathcal{N}(0,\sigma^2_{\epsilon})$ and is independent from the sampling location or the function value. We consider a discrete domain $\Omega$ with finitely many points where we would like to classify the function as either above the threshold or below it, with some degree of confidence. We assume that we are  allowed to query the function at very few points with noisy evaluations, in which case the unseen regions of the domain can only be classified with some probability, for example, $f(\mathbf x)>t$ with probability at least $\delta$. 
In the design of our algorithm, we assume that the function $f(\mathbf x)$ is a sample from a Gaussian process (GP). However, our asymptotic analysis in Section 4 is model independent, \textit{i.e.}, it holds without any additional probabilistic assumptions on the function measurements.
\label{sec:pre}
\subsection{Gaussian processes}\label{gp_prelim}

A GP $\{f(\mathbf x)\;|\;\mathbf x\in\Omega\}$ is a collection of random variables, any finite subset of which is distributed according to a multivariate Gaussian specified by the mean function $\mu(\mathbf x)$ and the kernel $k(\mathbf x,\mathbf x')$.
Suppose that we have a prior mean $\mu_0(\mathbf x)$ and kernel $k_0(\mathbf x,\mathbf x')$ for the GP, and $n$ (noisy) measurements $\{(\mathbf x_i,y_i)\}_{i=1}^n$, where $y_i=f(\mathbf x_i)+\epsilon_i$ and $\epsilon_i\sim N(0,\sigma_{\epsilon}^2)$ for $i=1,\dots,n$. The posterior of $\{f(\mathbf x)\;|\;\mathbf x\in\Omega\}$ is still a GP, and its mean and kernel functions can be computed analytically as follows:
\begin{equation}
\begin{split}
\label{eqn:GP}
&\mu_n(\mathbf x)=\mu_{\mathbf x_{1:n},y_{1:n}}({\mathbf x})=\mu_0(\mathbf x)+k_n(\mathbf x)^T(K_n+\sigma^2I)^{-1}(y_{1:n}-\mu_0(\mathbf x_{1:n})),\\
&k_n(\mathbf x,\mathbf x')=k_{\mathbf x_{1:n}}({\mathbf x},{\mathbf x'})=k_0(\mathbf x,\mathbf x')-k_n(\mathbf x)^T(K_n+\sigma^2I)^{-1}k_n(\mathbf x'),
\end{split}
\end{equation}
where $k_n(\mathbf x)=[k_0(\mathbf x,\mathbf x_1),\dots,k_0(\mathbf x,\mathbf x_n)]^T$, $K_n=[k_0(\mathbf x_i,\mathbf x_j)]_{i,j=1}^n$. In particular, the posterior variance at $\mathbf x$ is $\sigma_n^2(\mathbf x)=k_n(\mathbf x,\mathbf x)$. Intuitively, as the number of measurements increases, the actual $f$ will be gradually revealed.

\subsection{Framework for super-level set estimation}
Algorithm 1 is the conceptual framework of the algorithms for level set estimation, which is adopted in most, if not all of the related literature. 
\begin{algorithm}[h]
   \caption{\textbf{Framework for Estimating the Super-level Set}}
   \label{alg:Framwork}
\begin{algorithmic}
   \STATE {\bfseries Input:} prior mean $\mu_0$, kernel $k_0$, objective function $f$, threshold $t$, tolerance $\delta$.
   \FOR{$i=1, 2, \dots$}
   \STATE $\mathbf x^+ \leftarrow \textsc{SelectPoint}(GP)$
   \STATE Query the objective function at $ \mathbf x^+$ to obtain $y^+$
   \STATE Update $GP^+ \leftarrow GP$ using $(\mathbf x^+,y^+)$
   \ENDFOR
   \STATE  {\bfseries Output:} the estimated super-level set $I_{GP}:=\{\mathbf x\in\Omega\;|\;P_{GP}(f(\mathbf x)>t)>\delta\}$. 
\end{algorithmic}
\end{algorithm}
Here the last step in Algorithm 1 follows Eqn. (1), in which $P_{GP}$ is the probability measure defined according to the posterior Gaussian process (\textit{i.e.}, conditioned on the filtration $\{(\mathbf x_i,y_i)\}_{i=1,...,n}$).
The estimated super-level set is denoted $I_{GP}$.
We remark that one can also decide the membership of the estimated level set in an online fashion, as is done in \cite{Got13,Bog16}. 

\subsection{Notation and assumptions}
At timestep $n$, we have observed $\{(\mathbf x_i,y_i) \}_{i=1,...,n}$, where $\mathbf x_i$ is the $i$ th sampling location and $y_i$ is the resulting noisy observation.
We denote with the subscript $GP$ (\textit{e.g.}, $\mu_{GP}$, $\sigma_{GP}$ and $k_{GP}$) the quantities conditioned on the filtration $\{(\mathbf x_i,y_i) \}_{i=1,...,n}$. We use the subscript ${GP^+}$ to denote such quantities still conditioned on the filtration $\{(\mathbf x_i,y_i) \}_{i=1,...,n}$ and additionally on the sampling location denoted $\mathbf x^+$. Notice that, while $\mu_{GP}(\mathbf x)$ is a deterministic quantity that can be computed with the predictive Eqns. (\ref{eqn:GP}), $\mu_{GP^+}(\mathbf x)$ depends on the random outcome $y^+$ at $\mathbf x^+$ and is therefore a random variable.

Unless otherwise specified, we always restrict ourselves to a finite fixed grid ${\mathbf z_1},\dots,{\mathbf z_m}\in \Omega$ as the set of all candidate sampling locations in $\Omega$. Here $\mathbf{z}_{1:m}$ are all distinct. We will then slightly abuse notation to use $\Omega$ to denote the set of grid points $\mathbf{z}_{1:m}$, with $|\Omega|=m$. We also assume without loss of generality that the prior kernel $k_0$ is positive definite.

\section{Super-level set estimation}
This section begins with some remarks on the relation between super-level set estimation and binary classification, and then describes our \textsc{RMILE} algorithm.
\subsection{Relation to binary classification}
A GP uniquely specifies a probability distribution for the unseen examples $\mathbf x \in \Omega$ and can be used to infer the region where the threshold is exceeded with probability at least $\delta$.
At any point $\mathbf x$, the posterior distribution of the random variable $f(\mathbf x)$ is still normal \cite{GPML}. Let $\mu_{GP}(\mathbf x)$ and $\sigma^2_{GP}(\mathbf x)$ denote its posterior mean and variance, respectively. Then the condition $P(f(\mathbf x)>t)>\delta$ in Eqn. (1) can be reformulated as follows:
\begin{eqnarray}
\label{classificationrulebis}
\mu_{GP}(\mathbf x) - \beta  \sigma_{GP}(\mathbf x) >t, 
\end{eqnarray}
where $\beta$ is a fixed coefficient that depends on $\delta$. For a normal distribution, if $\delta = 97.5\%$ then $\beta \approx 1.96$.
The user is free to select $\delta$ according to the application. If human safety is at stake, $\delta$ should be sufficiently high to avoid misclassification, although this will also result in a smaller $I_{GP}$.

Define as $FP$ the set of \emph{false positives}, that is, the points $\mathbf x$ such that $f(\mathbf x) \leq t$ but are classified as $\in I_{GP}$, and $FN$ the set of \emph{false negatives}, which are all the points $\mathbf x$ such that $f(\mathbf x) > t$ but are classified as not in $ I_{GP}$.
The following is a straightforward observation:
\begin{lemma}
\label{lmm:classification}
The classification rule identified by  Eqn. (\ref{classificationrulebis}) minimizes the expected weighted misclassification error:
\begin{equation}
\E \left( \delta \mathbbm{1}\{\mathbf x \in \text{FP}\} + (1-\delta) \mathbbm{1}\{\mathbf x \in \text{FN}\} \right),
\end{equation}
among all deterministic classification rules under the posterior probability measure given by the Gaussian processes. Here, $\mathbf x\in \Omega$ is arbitrary and fixed.
\end{lemma}
In the above expression, the  expectation is conditioned on the filtration, \textit{i.e.}, on the observed samples $\{(\mathbf x_i,y_i)\}_{i=1,...,n}$.
We can see that rule (\ref{classificationrulebis}) penalize false positives much more than false negatives when $\delta$ is close to $1$. As a result, they are relatively conservative in the inclusion of points in the estimated supe-level set, and thus balance with our ``radical'' acquisition function to be introduced below.

\subsection{Robust maximum improvement in level-set estimation}
Our idea is to develop a method that aims to find the largest possible area where the function exceeds a given threshold with high probability. 

Let $I_{GP}$ be the set of points currently classified as above the threshold using the posterior GP. If we could sample at an arbitrary $\mathbf x^+$ and incorporate the feedback $y(\mathbf x^+) = f(\mathbf x^+) + \epsilon$ into the GP, then we would obtain a new Gaussian process, GP$^+$, which is a function of the (random) outcome $y(\mathbf x^+)$ and the sampling location $\mathbf x^+$. Then GP$^+$ can be used to infer an updated classification $I_{GP^+}$. 
We thus consider maximizing the volume of $I_{GP^+}$, \textit{i.e.}, $|I_{GP^+}| :=  \sum_{\mathbf x \in \Omega}  \mathbbm{1} \left\{ P_{GP^+}\left(f(\mathbf x)>t \right)> \delta \right\}$ to find the ``optimal'' sampling location. Equivalently, we would like to find the point $\mathbf x^+$ that would yield the maximum improvement  $|I_{GP^+}| -|I_{GP}| $ in expectation among all candidate sampling locations. Formally, the next sampling point is chosen as the solution to:
\begin{eqnarray}
\label{eqn:Lookahead}
\argmax_{\mathbf x^+\in\Omega}  \E_{y^+} \left|I_{GP^+}\right| - \left|I_{GP}\right|,
\end{eqnarray}
where the expectation is taken with respect to the random outcome $y^+$ (which is shorthand for $y(\mathbf x^+)$) resulting from sampling at $\mathbf x^+$, and is conditioned on the filtration $\{(\mathbf x_i,y_i)\}_{i=1,...,n}$. This criterion is similar to the expected improvement developed in the context of Bayesian optimization \cite{Mock78}, and is also closely related to the criteria of \cite{Ma14}, \cite{Ma15}. However, their framework differ from ours, and they all suffer from potential lack of exploration. In particular, \cite{Ma14}, \cite{Ma15} focus on region-level detection instead of point-wise detection, as we focus on here. Although Eqn. (\ref{eqn:Lookahead}) does fall as a special case of the acquisition function proposed in \cite{Ma15} when a single point is chosen as the ``linear functional'' there, their explanation for exploration inside each region becomes meaningless as each region then contains only a single point. As a result, no convergence guarantees have been established for these algorithms, despite their empirical success in certain problems. In particular, it remains an open issue how an intrinsic exploration strategy can be included. To this end, we modify criterion (\ref{eqn:Lookahead}) by introducing a trade-off between the posterior variances, which we prove to ensure certain asymptotic convergence, as we discuss below.

While Eqn. (\ref{eqn:Lookahead}) defines a reasonable acquisition function that seeks improvement in the discovery of points lying in the super-level set with high probability, it may suffer from potential model misspecification and lack of exploration. Consider an extreme case when all the points are classified as above the threshold by the chosen prior at the beginning. Then Eqn. (\ref{eqn:Lookahead}) may lead the procedure to stall and repeatedly sample at locations with the largest function values to maintain the largest super-level set specified by the prior.  To remedy this issue, we modify the criterion in Eqn. (\ref{eqn:Lookahead}) so that the algorithm cannot ``get stuck'' indefinitely. To achieve this goal, we guarantee a minimum positive exploration bonus everywhere by introducing a marginal variance term. Let $|I_{GP}^\epsilon| =  \sum_{\mathbf x \in \Omega}  \mathbbm{1} \left\{ P_{GP}\left(f(\mathbf x)>t - \epsilon \right)> \delta \right\}$ for $\epsilon > 0$, which is essentially a shift of the threshold from $t$ to $t-\epsilon$ (the shift is mostly for technical reasons to simplify the analysis).  Our final \emph{acquisition function} is then defined as:
\begin{eqnarray}
\label{eqn:Lookahead_Convergence}
E_{GP}(\mathbf x^+) = \max\left\{\E_{y^+} \left|I_{GP^+}\right| - \left|I_{GP}^\epsilon \right| , \gamma \sigma_{GP}(\mathbf x^+)\right\},
\end{eqnarray}
for some small constants $\epsilon > 0, \gamma > 0$. Intuitively, the additional variance term ensures that the algorithm moves to a region with a higher variance when the expected improvement is sufficiently reduced at the current point.

\subsection{Efficient implementation}
At each candidate sample point $\mathbf x^+\in\Omega$, to evaluate Eqn. (\ref{eqn:Lookahead_Convergence}), we would need to sample $f(\mathbf x^+)$ from the current GP, from which we can compute the posterior classification and repeat this procedure in a Monte Carlo fashion to estimate the expected improvement.

Nevertheless, it is possible to avoid sampling from the GP here. To see this, notice that by Eqns. (2), the variance $\sigma^2_{GP^+}(\mathbf x)$ is unaffected by a new observation $y^+$ as it only depends on the sampling location $\mathbf x^+$. On the other hand, the posterior mean $\mu_{GP^+}(\mathbf x)$ is linearly correlated with the sample $y^+$ (to indicate this dependency we would rewrite it as $\mu_{GP^+}(\mathbf x;y^+)$). Therefore, it is possible to compute the  outcome $y^+$ that would change the classification for point $\mathbf x$ under consideration, that is, we only need to determine the ``limit'' value for the new sample $y^+$ that turns the indicator $\mathbbm{1} \left\{ P_{GP^+}\left(f(\mathbf x)>t \right)> \delta \right\}$ on or off in the computation of $\E_{y^+}|I_{GP^+}|$. As a result, we obtain the following expression:
\begin{lemma}\label{fast_compute}
$ \E_{y^+} \left|I_{GP^+}\right|$ obtained by sampling at $\mathbf x^+$ can be computed analytically as follows:
\begin{equation}
\sum_{\mathbf x \in \Omega} \Phi \left(\frac{\sqrt{\sigma^2_{GP}(\mathbf x^+) + \sigma^2_{\epsilon}
 }}{|\textup{Cov}_{GP}(f(\mathbf x), f(\mathbf x^+))|} \times \left( \mu_{GP}(\mathbf x) - \beta\sigma_{GP^+}(\mathbf x) -t  \right) \right) 
\label{objective}
\end{equation}
where $\Phi(\cdot)$ is the cumulative distribution function (CDF) of the standard normal random variables, and 
\begin{equation}
\sigma^2_{GP^+}(\mathbf x)=\textup{Cov}_{GP^+}(f(\mathbf x),f(\mathbf x)) = \sigma^2_{GP}(\mathbf x) - \dfrac{\textup{Cov}_{GP}^2(f(\mathbf x), f(\mathbf x^+))}{\sigma^2_{GP}(\mathbf x^+) + \sigma^2_{\epsilon}},
\end{equation}
and $\textup{Cov}_{GP}(f(\mathbf x), f(\mathbf x^+))=k_{GP}(\mathbf x,\mathbf x^+)$ is the (current) posterior covariance between $f(\mathbf x)$ and $f(\mathbf x^+)$.
\end{lemma}
In the above derivation, we are implicitly assuming that $f(\mathbf x)$ can be modeled as a sample from the GP. The posterior covariance $\textup{Cov}_{GP}(f(\mathbf x), f(\mathbf x^+))$ can be calculated, for example, by using Eqns. (\ref{eqn:GP}). For a fixed grid, such computation can be rearranged so that the full posterior covariance matrix is stored and updated at each iteration through rank-one updates. Different trade-offs between computational and memory complexity are also possible. Lemma \ref{fast_compute} is also shown in \cite{Ma14}, \cite{Ma15}, but to reduce notational confusion in cross-referencing, we provide a different and more direct proof in the appendix.

The \textsc{RMILE} algorithm follows the super-level set estimation framework described in Algorithm \ref{alg:Framwork}. At each iteration, it calls \textsc{SelectPoint} (Algorithm \ref{alg:ConceptualLookahead}). For a fixed sampling location $\mathbf x^+$, the algorithm computes the acquisition function (\ref{eqn:Lookahead_Convergence}) using Eqn. (\ref{objective}). 

Although it is possible to identify the level set at any time during the execution, we do not enforce an online classification scheme as in \cite{Got13,Bog16}. Instead, the classification is done offline using all  available information, which makes the algorithm work better in practice, as can be seen in the numerical experiments.

\begin{algorithm}[H]
   \caption{\textbf{\textsc{RMILE} (\emph{SelectPoint(GP)})}}
  \label{alg:ConceptualLookahead} 
\begin{algorithmic}
   \STATE {\bfseries Input:} Current GP, constants $\gamma>0$, $\epsilon>0$.
   \FOR{$\mathbf{x^+} \in \Omega$}
   \STATE Compute $E_{GP}(\mathbf x^+) = \max\{\E_{y^+} \left|I_{GP^+}\right| - \left|I_{GP}^\epsilon \right| , \gamma \sigma_{GP}(\mathbf x^+)\}$
   \ENDFOR
   \STATE {\bfseries Output:} $\mathbf x^*= \argmax_{\mathbf x^+\in\Omega} E_{GP}(\mathbf x^+)$.
\end{algorithmic}
\end{algorithm}

\subsection{Connection to uncertainty/variance reduction}
So far we have assumed that the threshold is known a priori. In fact, the design process in engineering typically involves several ``iterations'' where the conceptual idea is revised, leading to changes in the requirements or the threshold. In such cases, one would like to obtain a model of the function that can later be used to identify different regions, say $I^{(t_1)}_{GP},I^{(t_2)}_{GP}, I^{(t_3)}_{GP}$ corresponding to different thresholds $t_1,t_2,t_3$ (this notation should not be confused with $I_{GP}^{\epsilon}$ previously used to indicate a shift in the threshold).  For three thresholds, this can be easily done by redefining the objective function to maximize:
\begin{equation}
\E_{y^+}\left(|I^{(t_1)}_{GP^+}|+|I^{(t_2)}_{GP^+}|+|I^{(t_3)}_{GP^+}|\right) - \left(|I^{(t_1-\epsilon)}_{GP}|+|I^{(t_2-\epsilon)}_{GP}|+|I^{(t_3-\epsilon)}_{GP}|\right),
\label{multithresh}
\end{equation}
so that the algorithm is biased towards identifying all three of them. If, for example, $f(\mathbf x)>\text{max}(t_1,t_2,t_3)$, then that point will contribute three times as much to the objective function. 

We can naturally extend the idea of Eqn. (\ref{multithresh}) and look for all thresholds in a given range $[a,b]$. That is, the finite summation in Eqn. (\ref{multithresh}) can be replaced by an integral over all possible thresholds:  $\E_{y^+}\int_{-\infty}^{\infty} |I^{(t)}_{GP^+}|-|I_{GP}^{(t-\epsilon)}| dt$.  Interestingly, if one considers the extreme case when $\epsilon=0$, then our algorithm reduces to a type of variance minimization:
\begin{lemma}
\label{lemma:varmin}
If the acquisition function is redefined as:
\begin{equation}\label{eqn:varred}
\begin{split}
E_{GP}^{var}(\mathbf x^+):=&\E_{y^+}\int_{-\infty}^{\infty} |I^{(t)}_{GP^+}| - |I^{(t)}_{GP}| dt,
\end{split}
\end{equation}
then Algorithm 2 minimizes the $l_1$-norm of the posterior standard deviation, \textit{i.e.}, the next query point $\mathbf x^+$ is selected as $\mathbf x^+=\argmin_{\mathbf x^+ \in \Omega} \sum_{\mathbf x \in \Omega} \sigma_{GP^+}(\mathbf x)$.
\end{lemma}
Since we can recast the objective function as $\sum_{\mathbf x \in \Omega} \sigma_{GP^+}(\mathbf x) -\sigma_{GP}(\mathbf x)$, the acquisition function (\ref{eqn:varred}) chooses the point that maximizes the reduction in the standard deviation across the domain.
This is similar to the acquisition function used in \cite{Bog16} for an appropriate choice of the parameters.

\section{Asymptotic behavior on finite grids}
In the absence of noise, a well-designed algorithm should avoid re-sampling at the same location since additional information is not acquired. In other words, on finite grids every point should be sampled at most once before an algorithm terminates.

In the case of noisy measurements, however, an algorithm may need to re-sample at the same location multiple times in order to get a more accurate estimate of the function value.
Intuitively, so long as an algorithm samples each point of the grid infinitely often, the underlying function should be gradually revealed \cite{GPML}. Below, we first validate some reasonable assumptions that guarantee the asymptotic convergence of a generic acquisition function in Algorithm \ref{alg:Framwork}, and then show that \textsc{RMILE} satisfies these conditions.

\begin{lemma}
\label{lemma:convergence}
Let $E_{GP}(\mathbf x^+)$ be an acquisition function that depends on the posterior $GP$ at a potential query point $\mathbf x^+$, such that for sufficiently small $\sigma^2_{GP}(\mathbf x^+)$, there exists a function $u(\cdot)$ which only depends on the posterior variance $\sigma^2_{GP}(\mathbf x^+)$, with
\begin{equation}
\label{eqn:bracket}
E_{GP}(\mathbf x^+) \leq u(\sigma_{GP}(\mathbf x^+)),\quad \lim_{\sigma(\mathbf x^+)\to 0^+} u(\sigma_{GP}(\mathbf x^+)) = 0.
\end{equation}
In addition, assume that there exists a global lower bound $l(\cdot)$, such that 
\begin{equation} 
E_{GP}(\mathbf x^+)\geq l(\sigma_{GP}(\mathbf x^+)),\quad \lim_{\sigma(\mathbf x^+)\to 0^+} l(\sigma_{GP}(\mathbf x^+)) = 0,
\end{equation}
and assume that $l(\sigma_{GP}(\mathbf x^+))$ is strictly increasing in $\sigma_{GP}(\mathbf x^+)$.

If Algorithm \ref{alg:Framwork} selects the next query point as $\argmax_{\mathbf x^+} E_{GP}(\mathbf x^+) $ and is run without termination, then there cannot be a point in the grid that is sampled only finitely many times.
\end{lemma}
The lemma does not assume that the true function can be represented as a sample from a Gaussian process; only the upper and lower bounds on the acquisition function are needed as a function of the posterior marginal variance.
The intuition is that $E_{GP}(\mathbf x^+)$ can fluctuate as the sampling process progresses; however, as the variance $\sigma_{GP}(\mathbf x^+)$ of a point is progressively reduced, one more sample at $\mathbf x^+$ should bring less and less improvement as measured by $E_{GP}(\mathbf x^+)$. This implies that the algorithm will move to a location where $E_{GP}(\cdot)$ is higher. The proof can be found in the appendix.

We are now ready to verify the robustness of our algorithm: as we show next, it satisfies the assumption of Lemma \ref{lemma:convergence}. Let  $\overline{\sigma}^2:=\max_{i=1,\dots,m}\sigma_{0}^2(\mathbf z_i)$, where $\mathbf z_i$ are the grid points in $\Omega$.
\begin{lemma}
\label{lemma:Lookahead_convergence}
For the acquisition function (\ref{eqn:Lookahead_Convergence}) with $\gamma > 0$, $\epsilon > 0$, we have:
\begin{itemize}
\item $u(\sigma_{GP}(\mathbf x^+)) = \max \left( |\Omega| \Phi \left(\frac{\sigma_{\epsilon}}{\bar{\sigma} \sigma_{GP}(\mathbf x^+)} \left(-\epsilon/2 \right) \right),\gamma \sigma_{GP}(\mathbf x^+) \right)$ 
\item $l(\sigma_{GP}(\mathbf x^+)) = \gamma \sigma_{GP}(\mathbf x^+)$ 
\end{itemize}
Also the lower bound $ l(\sigma_{GP}(\mathbf x^+))$ is monotonically increasing and 
$$\lim_{\sigma(\mathbf x^+)\to 0^+} u(\sigma(\mathbf x^+)) = \lim_{\sigma(\mathbf x^+)\to 0^+} l(\sigma(\mathbf x^+)) = 0.$$
\end{lemma}
The roles of $\epsilon$ and $\gamma$ are important  in terms of the asymptotic behavior. More precisely, the modification that leads to Eqn. (\ref{eqn:Lookahead_Convergence}) ensures a minimum exploration bonus given by $\gamma \sigma_{GP}(\mathbf x^+)$ and is a crucial difference compared to \cite{Ma14}, \cite{Ma15}.

\section{Numerical experiments}
This section empirically assesses the proposed procedure on numerical experiments. We use a standard squared exponential kernel 
\begin{equation}
k(\mathbf x, \mathbf x') = \sigma^2_{ker} \exp(-\| \mathbf x - \mathbf x'\|_2^2/(2l^2))
\end{equation}

We start by examining the effectiveness of the robust modifications, and then proceed to comparing our proposed approach with state-of-the art algorithms. Although in principle the model noise level $\sigma_{\epsilon}$ can be different from the algorithm noise level (which we also denote as $\sigma_{\epsilon}$ with slight abuse of notation), we typically take them to be the same as is done conventionally in the literature, unless otherwise stated (\textit{e.g.}, in the next subsection). We also emphasize that to make the comparisons fair, the performance of all the algorithms is evaluated with classification criterion (\ref{classificationrulebis}), instead of the criteria proposed in the original papers (\textit{e.g.}, posterior mean).

\subsection{Robustification effects}
This section shows how the robust adjustment parameters $\epsilon$ and $\gamma$ in Eqn. (\ref{eqn:Lookahead_Convergence}) help improve the performance of the algorithm. We compare two sets of parameters: a) $\epsilon=\gamma= \num{1e-8}$; b) $\epsilon=0$, $\gamma=-\infty$. Notice that with parameter set b), Eqn. (\ref{eqn:Lookahead_Convergence}) reduces to Eqn. (\ref{eqn:Lookahead}), \textit{i.e.}, the one without guaranteed convergence.

To stabilize the performance, we first sample at 3 points chosen uniformly at random as seeds, and compute the resulting posterior distribution as the prior. To showcase the robustness of our algorithm, we keep the prior mean to be $0$, and $\ln(\sigma_{ker})=4$, $\ln(l)=1$ throughout the experiments in this subsection. For each problem, we run $25$ simulations of our algorithm. 

\begin{figure*}[h]
\centering
\includegraphics[width=5.4cm]{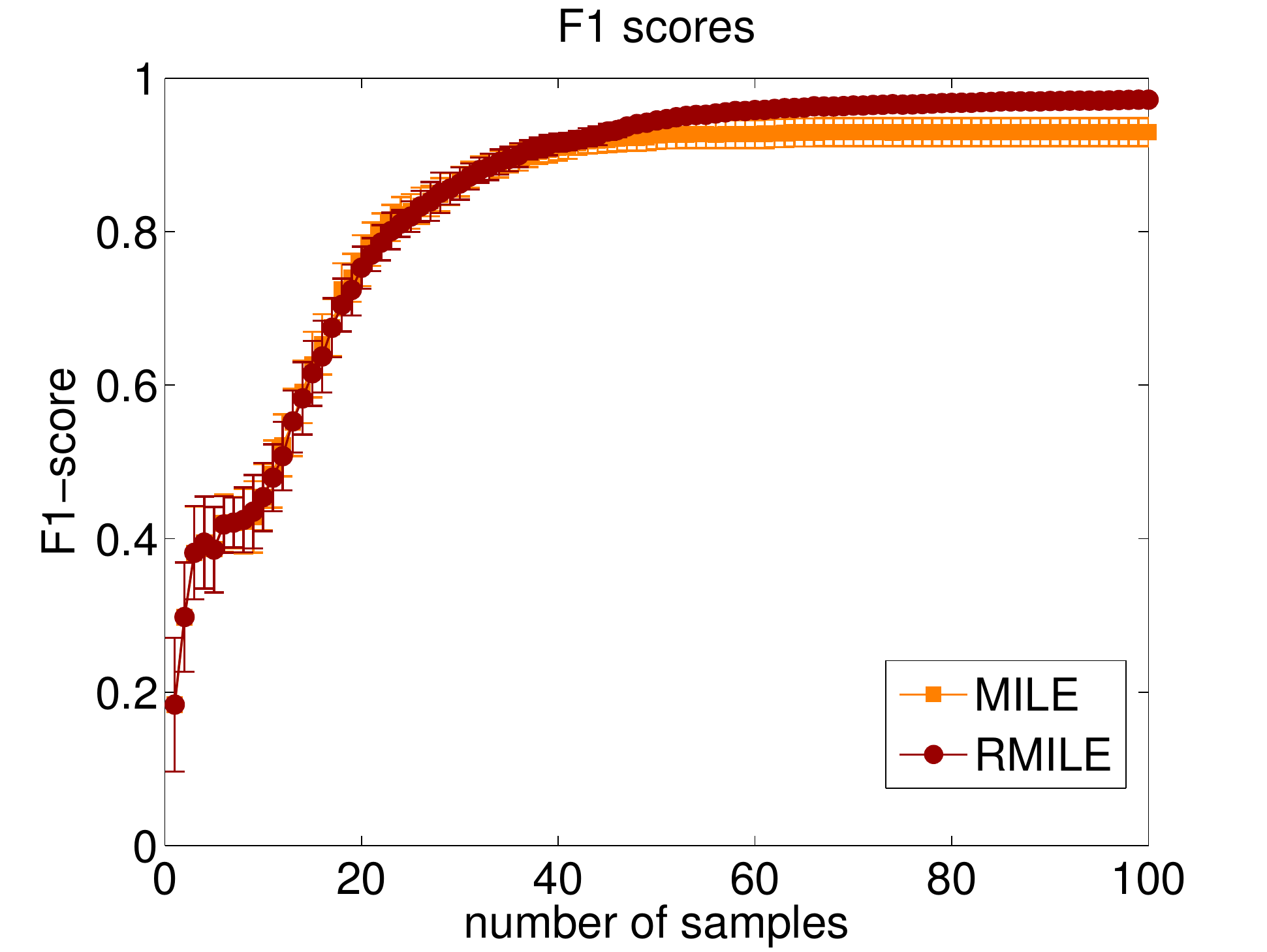}
\includegraphics[width=5.4cm]{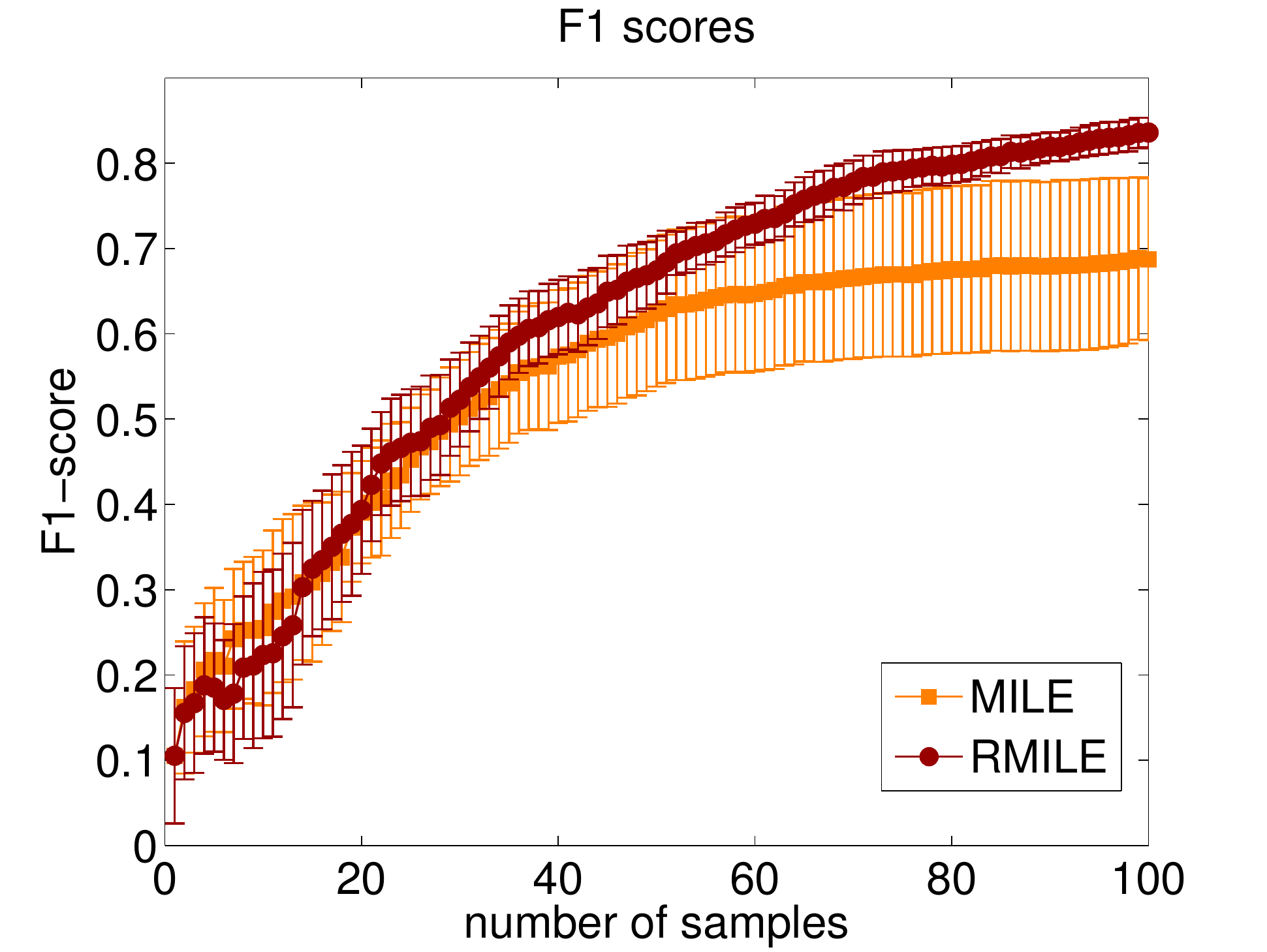}
\caption{Himmelblau's function. Left: small noise. Right: large misspecified noise.}
\label{robust_himm}
\end{figure*}

We consider the negative Himmelblau's function (Figure \ref{fig:himplot}) defined in $[-5,5]\times[-5,5]$, a commonly used multi-modal function. We take a uniform grid of $30\times 30$ points, and the threshold is set to $t=-50$. Here we consider two sets of noise levels: 1) a small noise setting with both model and algorithm noise levels $\sigma_{\epsilon}=0.1$; 2) a misspecified large noise setting, with model noise level $30$ and algorithm noise level $3$. The results are shown in Figure \ref{robust_himm}. Here we label parameter set a) as ``\textsc{RMILE}'' and parameter set b) as ``\textsc{MILE}''. We can see that in both cases, the robust version outperforms the vanilla one, and the difference is more dramatic in the second (harder) case.

Our algorithm is quite robust to the parameter choices of $\epsilon$ and $\gamma$, so long as they are positive. In particular, we obtained almost the same performance  as above when setting $\epsilon=\gamma= \num{1e-2}$.

For simplicity, hereafter we set $\epsilon = 10^{-12}$ and $\gamma = 10^{-10}$. We compare our approach against the Straddle heuristic \cite{Bry05} and the \textsc{LSE} algorithm \cite{Got13}, which are the most relevant algorithms to our work. Another relevant approach is the \textsc{TruVaR} algorithm, which has been found to perform similarly to \textsc{LSE} in numerical experiments for level-set estimation \cite{Bog16}. 

\subsection{2D synthetic examples}

\begin{figure*}[h]
\centering
\includegraphics[width=4.3cm]{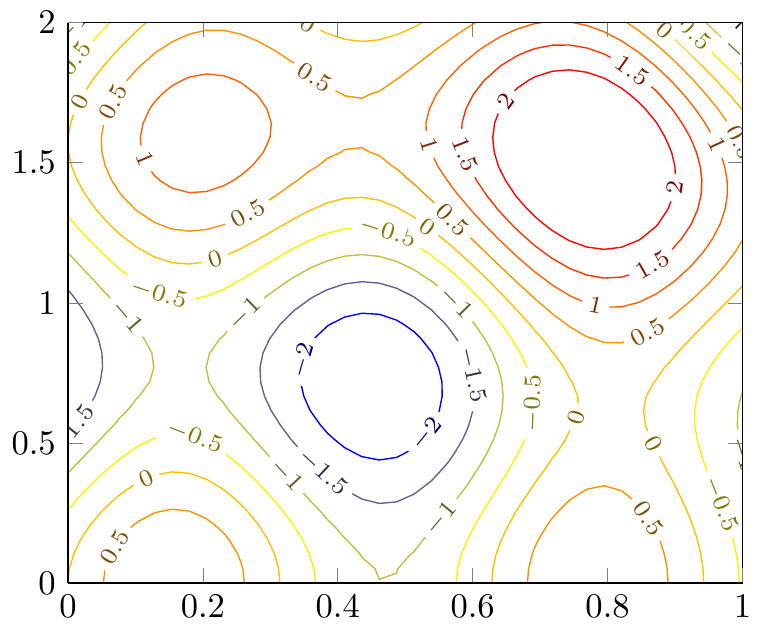}
\includegraphics[width=4.3cm]{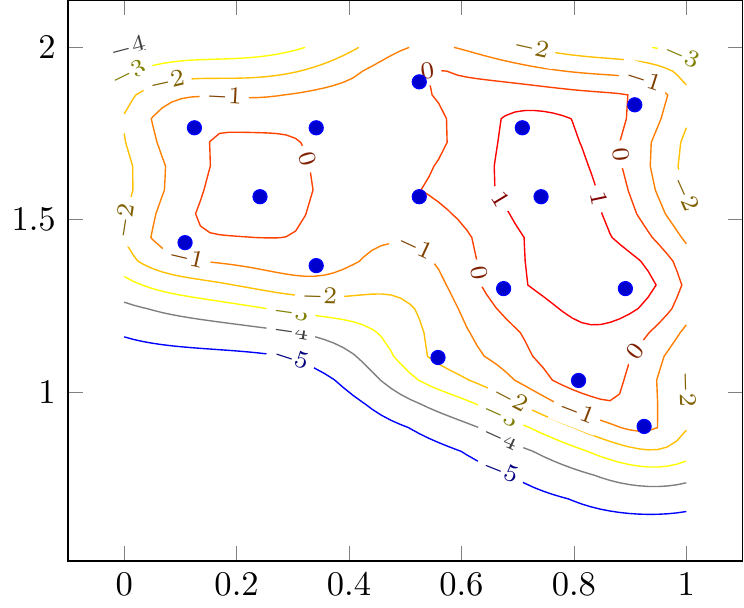}
\includegraphics[width=4.3cm]{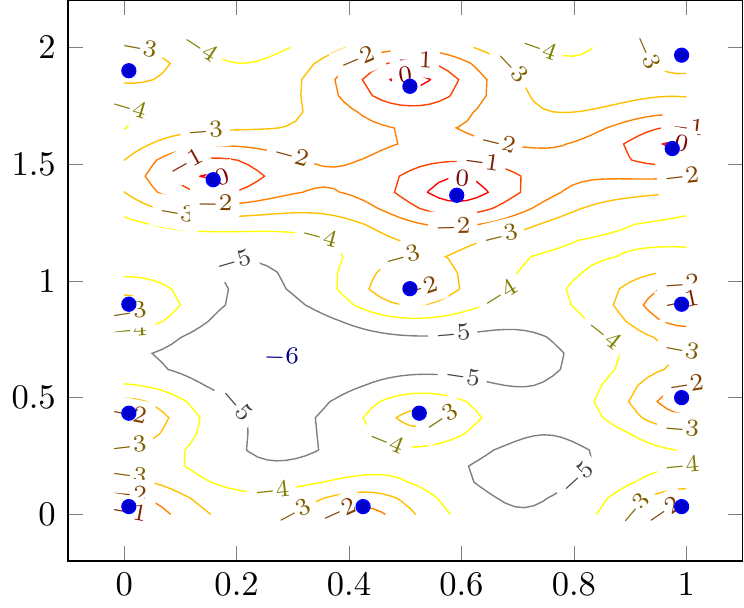}
\includegraphics[width=4.3cm]{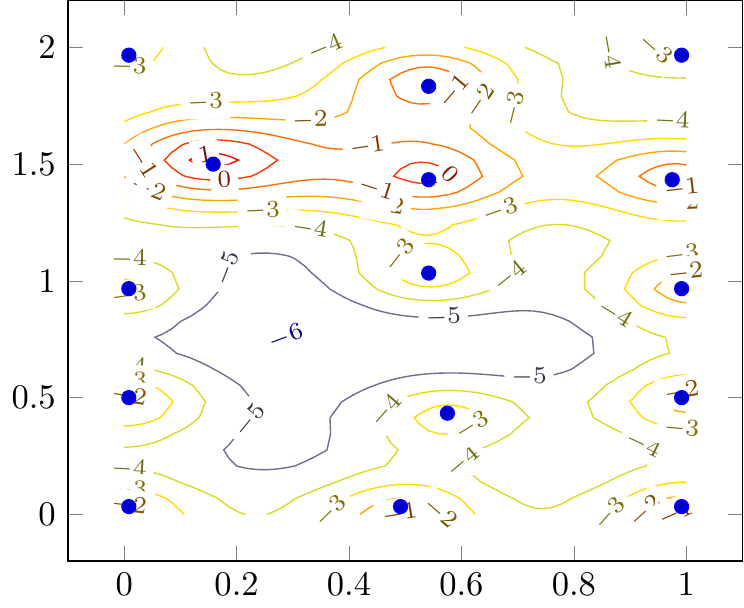}
\caption{Top left: true contours of the sinusoidal function. Location of the first 15 samples along with the  contours given by the GP for $\mu_{GP}(\mathbf x) - 1.96\sigma_{GP}(\mathbf x)$ for \textsc{RMILE} (top right), Straddle (bottom left) and \textsc{LSE} (bottom right).}
\label{fig:sineplot}
\end{figure*}

We consider a sinusoidal function $\sin(10x_1) + \cos(4x_2) - \cos(3x_1x_2)$, whose contours are plotted in Figure \ref{fig:sineplot} defined in the box $[0,1]\times[0,2]$. We superimpose a grid of $30 \times 60$ points uniformly separated and run 25 simulations for our algorithm \textsc{RMILE} along with \textsc{LSE} and the Straddle heuristic. The normally distributed noise has standard deviation $\ln(\sigma_{\epsilon}) = -1$, and the prior has uniform mean $= 0$ with $\ln(\sigma_{ker}) = 1.0 $ and $\ln(l) = -1.5$.  The threshold is set to $t=1$.
These are supposed to be representative of the prior knowledge that the user may have about the function at hand, but are not necessarily the hyper-parameters that maximize the likelihood of some held-out data under the Gaussian process. 

Similarly, we also consider again the Himmelblau's function. We run $25$ simulations on a $50 \times 50$ grid for our algorithm, the \textsc{LSE} algorithm and the Straddle heuristic. We assume that the true Himmelblau's function can be evaluated with some normally distributed noise with standard deviation $\ln(\sigma_{\epsilon}) = 2.0$ and mean zero. The threshold chosen for the experiment is $t = -100$, with a prior mean of $-100$, prior standard deviation $\ln(\sigma_{ker}) = 4$, and $\ln(l) = 0 $. 

\begin{figure*}[h]
\centering
\includegraphics[width=4.3cm]{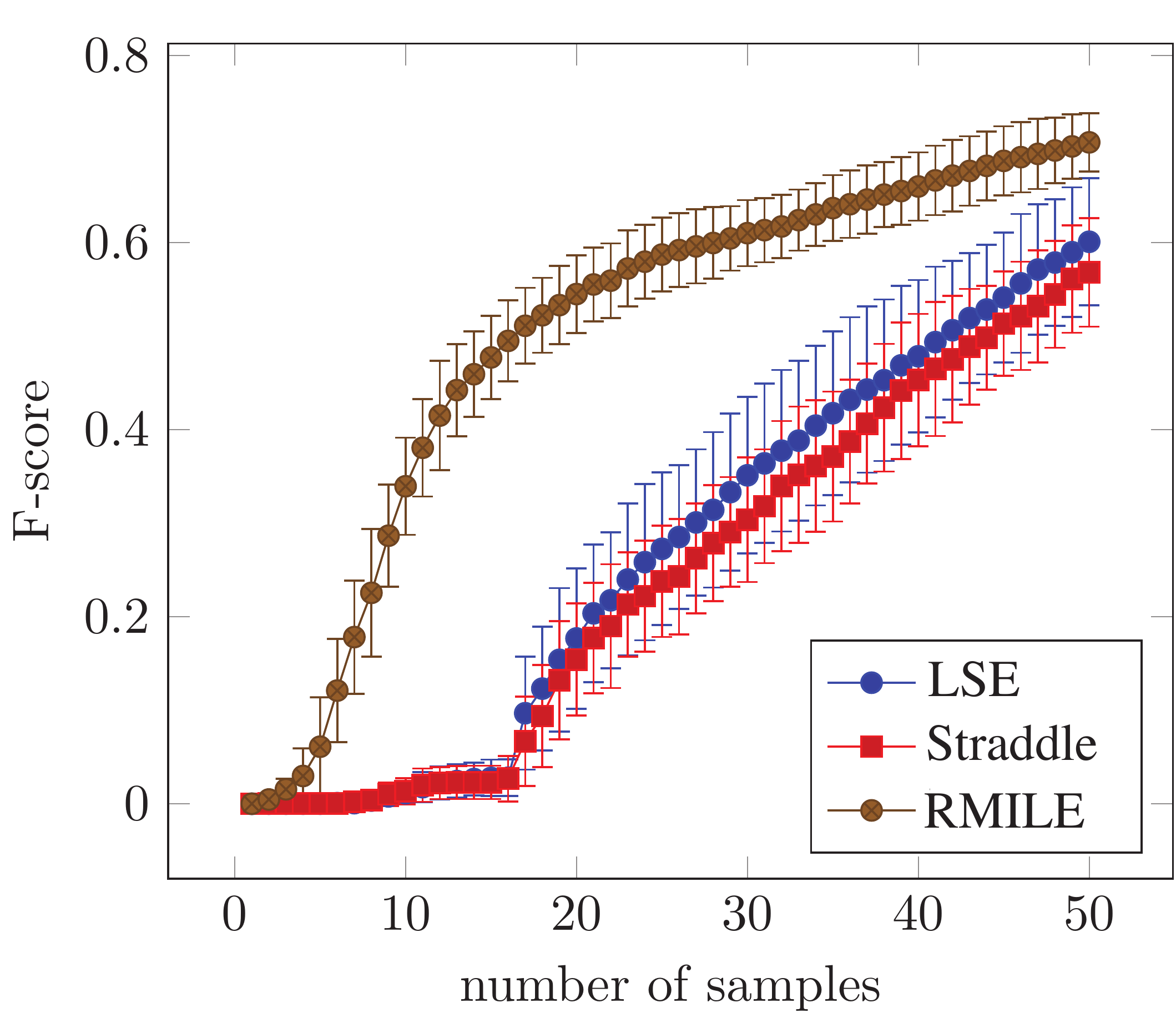}
\includegraphics[width=4.3cm]{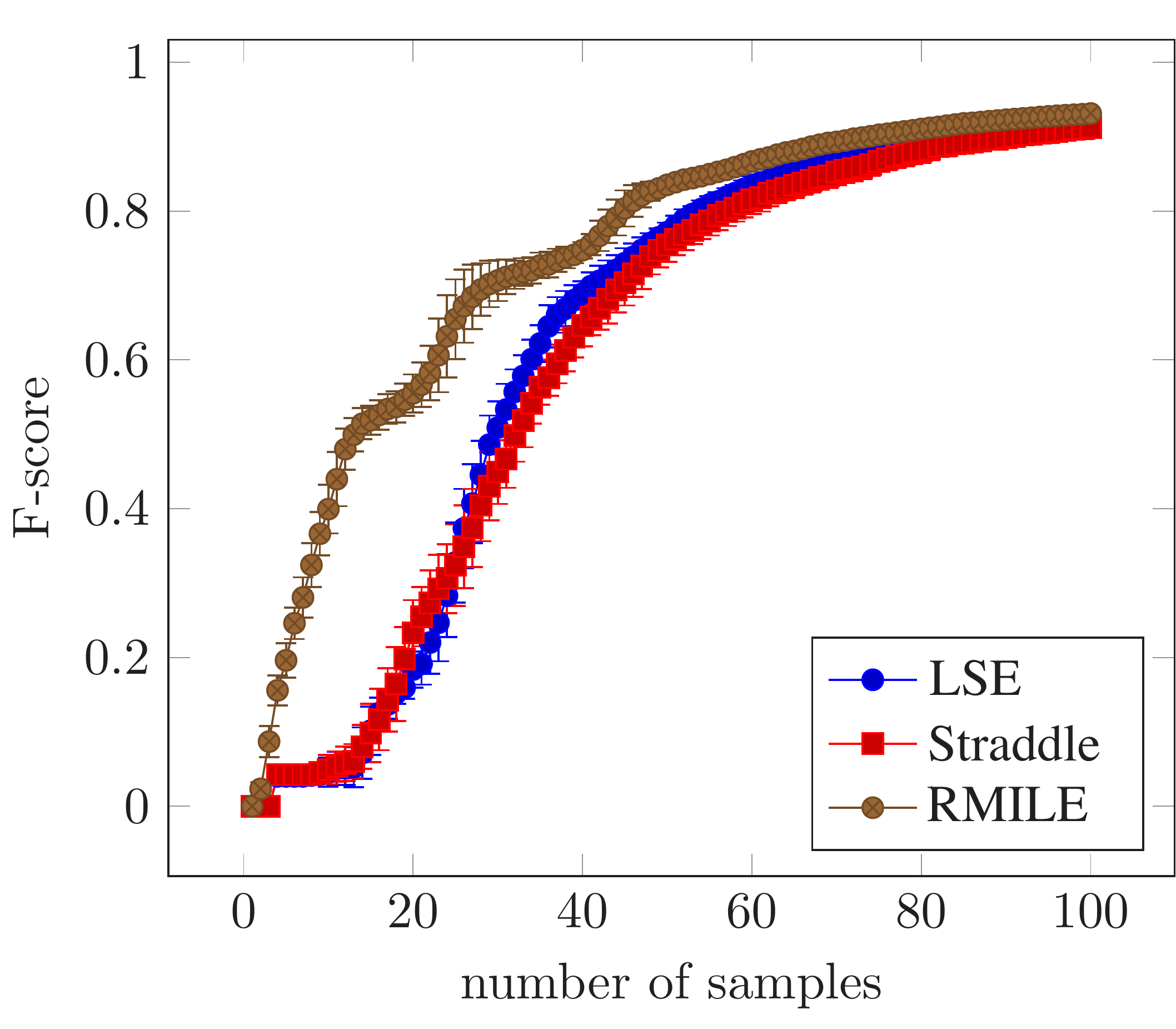}
\includegraphics[width=4.3cm]{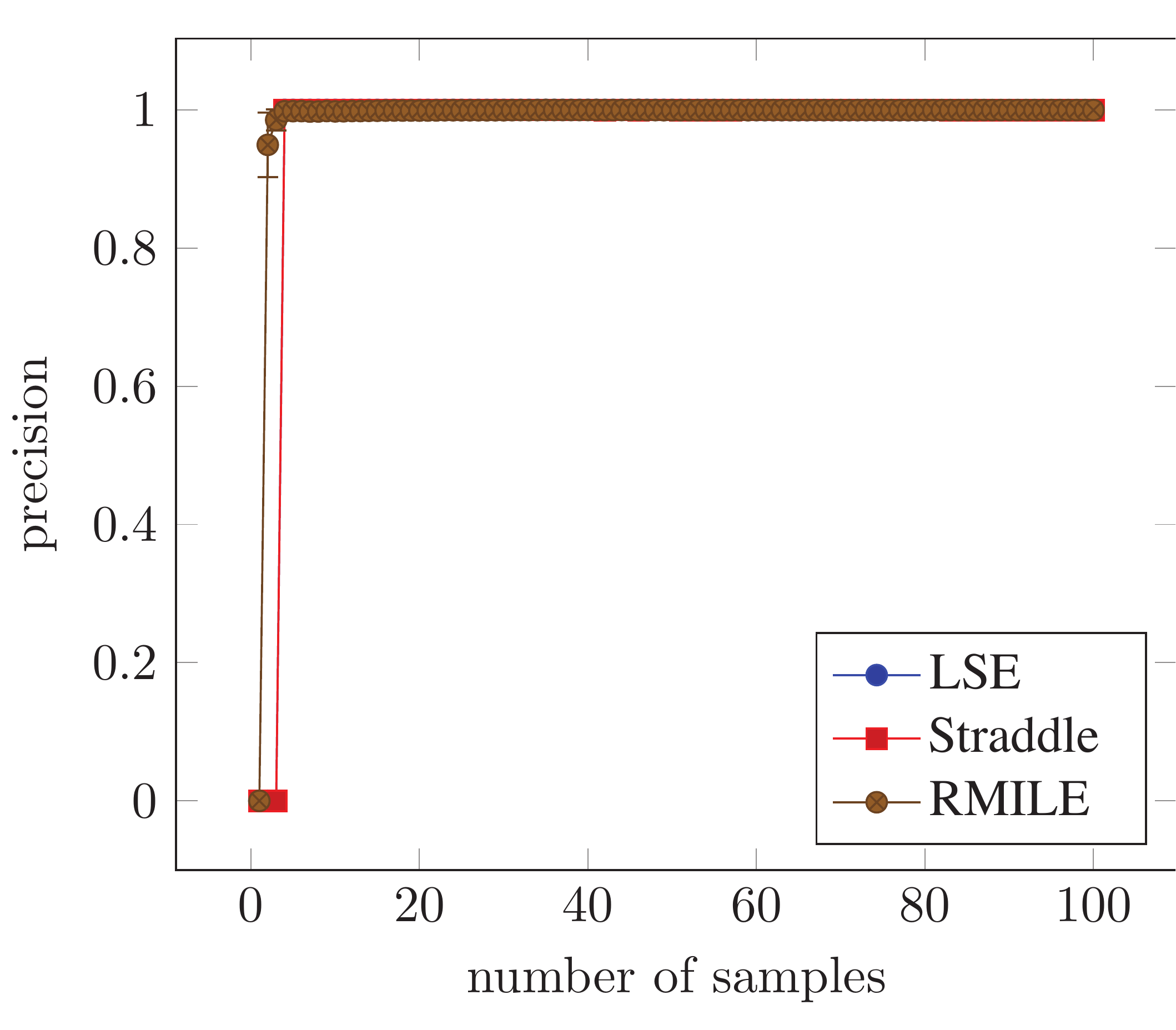}
\includegraphics[width=4.3cm]{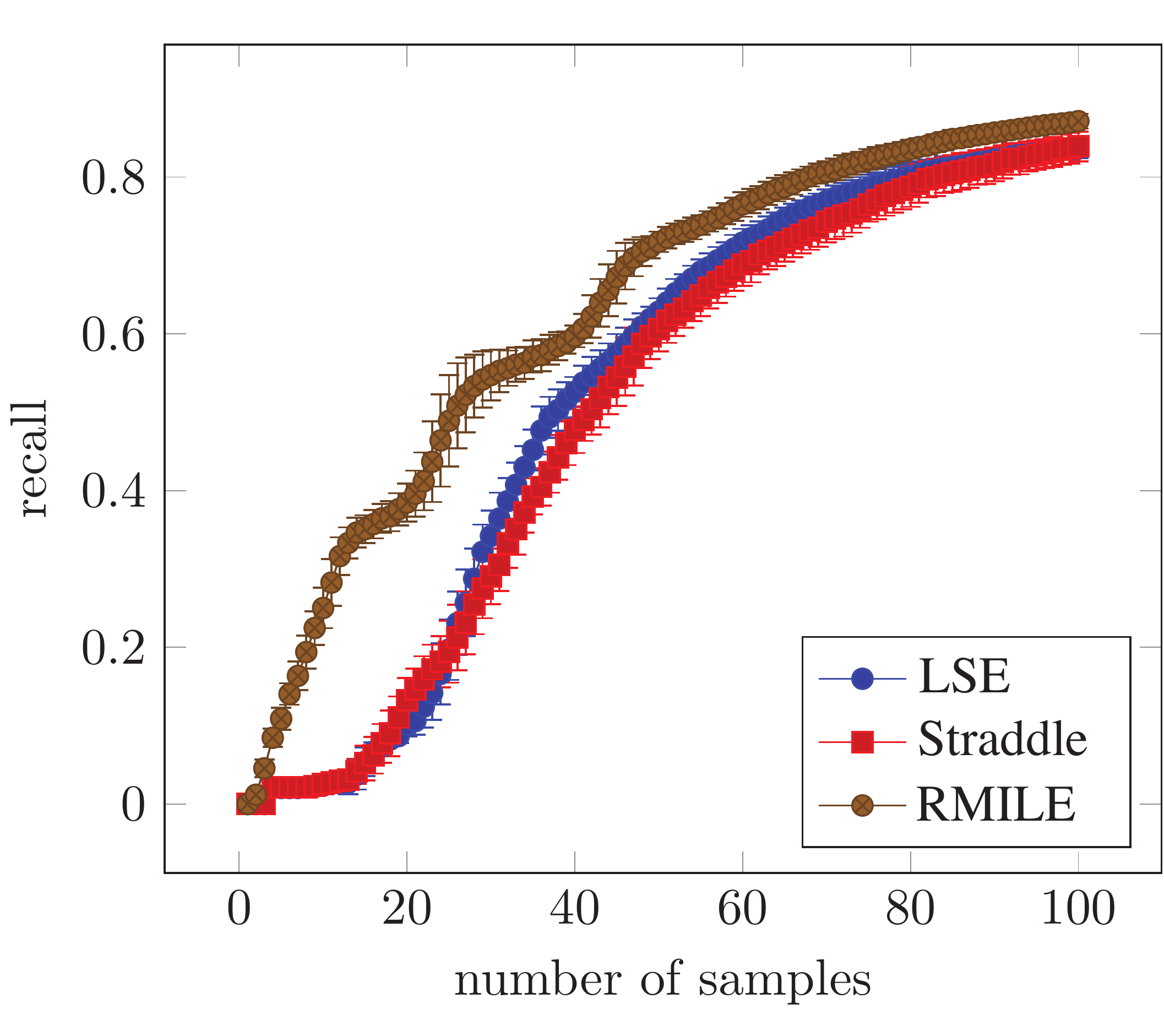}
\caption{$F_1$-score on the sinusoidal function (top left), and Himmelblau's function (top right). The means and confidence intervals of precision (bottom left) and recall (bottom right) refer to the Himmelblau's function experiment.}
\label{fig:score}
\end{figure*}

The advantage of the procedure with respect to the state of the art is demonstrated by the $F_1$-score on the sinusoidal and Himmelblau's function (Figure \ref{fig:score}).
We also show precision and recall separately for the numerical experiments on Himmelblau's function in Figure \ref{fig:score} bottom left and bottom right, respectively. 
 
\begin{figure*}[h]
\centering
\includegraphics[width=4.3cm]{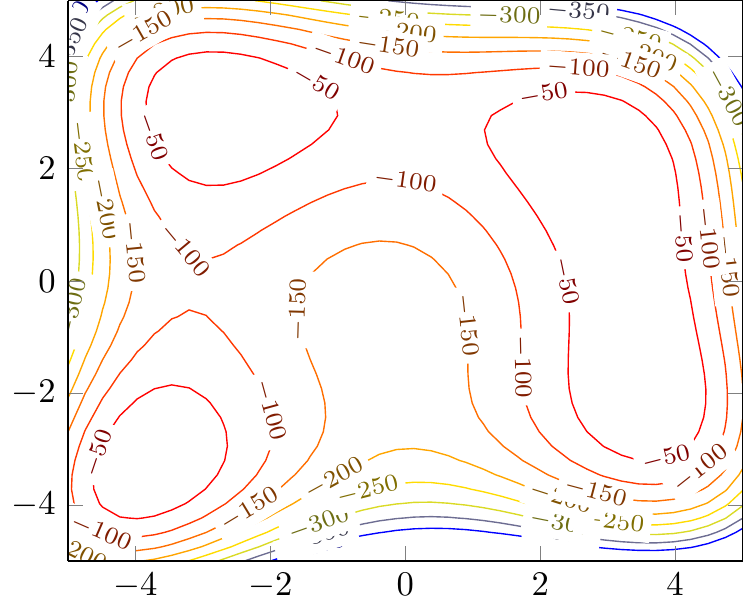}
\includegraphics[width=4.3cm]{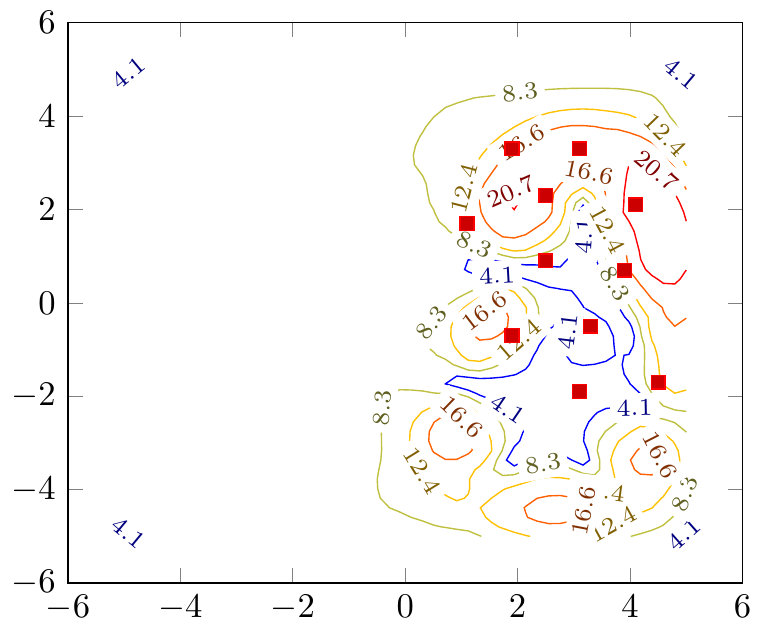}
\includegraphics[width=4.3cm]{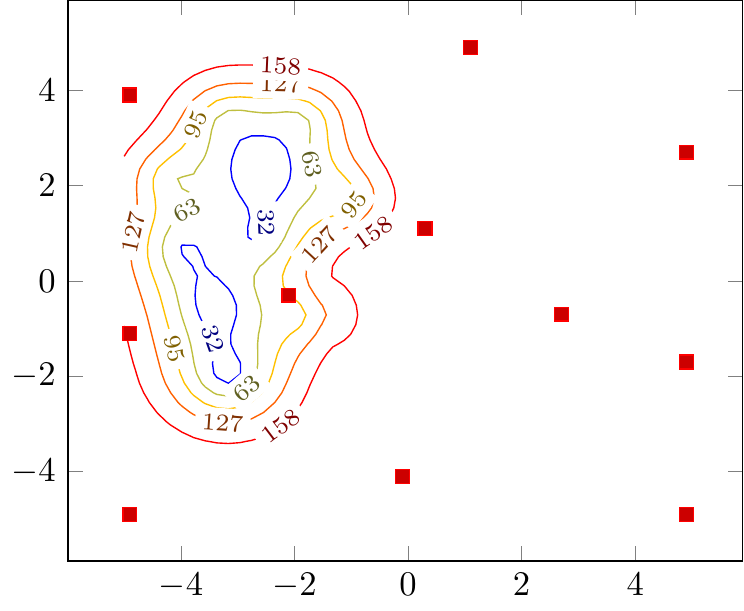}
\includegraphics[width=4.3cm]{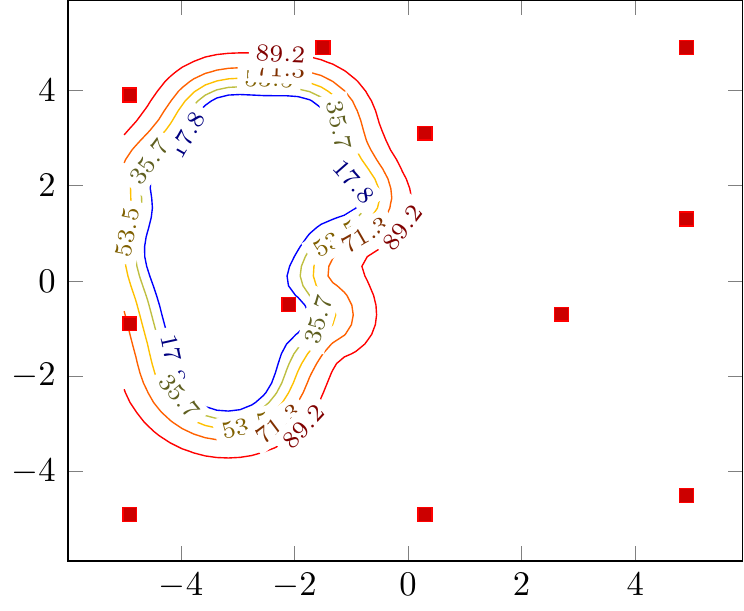}
\caption{Top left: true contours of the Himmelblau's function. Top right: value of the expected improvement for \textsc{RMILE}. Bottom left: Straddle score. Bottom right: ambiguity for the \textsc{LSE} algorithm. The locations of the first 11 samples for the first run are superimposed.}
\label{fig:himplot}
\end{figure*}

In both numerical experiments it is relatively easy to find an initial point above the threshold. Our algorithm then proceeds by expanding $I_{GP}$ as much as possible at each step (Figure \ref{fig:sineplot}). This is in contrast to the Straddle heuristic and \textsc{LSE} which seek to reduce the variance and thus tend to sample more widely  in the initial phase. Notice that Straddle and \textsc{LSE} maximize a similar objective function for the selection of the next point, the ``Straddle score'' and the ``ambiguity'', respectively. However, at least in the initial exploration phase, these metrics have fairly uniform high value (Figure \ref{fig:himplot}) across the domain because the variance given by the Gaussian process is initially high. In contrast, \text{RMILE} gives higher scores to points in $\Omega$ that are likely to improve $I_{GP}$ the most, and therefore chooses to expand the current region above the threshold as much as possible before exploring regions far away from the current samples. 

Thus, our algorithm is more suitable especially when a very limited exploration budget is available and one cannot afford to reconstruct a good model of the function.
Although we use the well-established $F_1$-score for comparison, it may not always be the most appropriate and fair metric for our proposed problem. In particular, as we noted in Lemma \ref{lmm:classification}, our classification rule penalizes false positives far more than false negatives.

\subsection{Simulation experiments: aircraft collision avoidance}

\begin{figure*}[h]
\centering
\includegraphics[width=3.9cm,trim={3cm 0 3cm 0},clip]{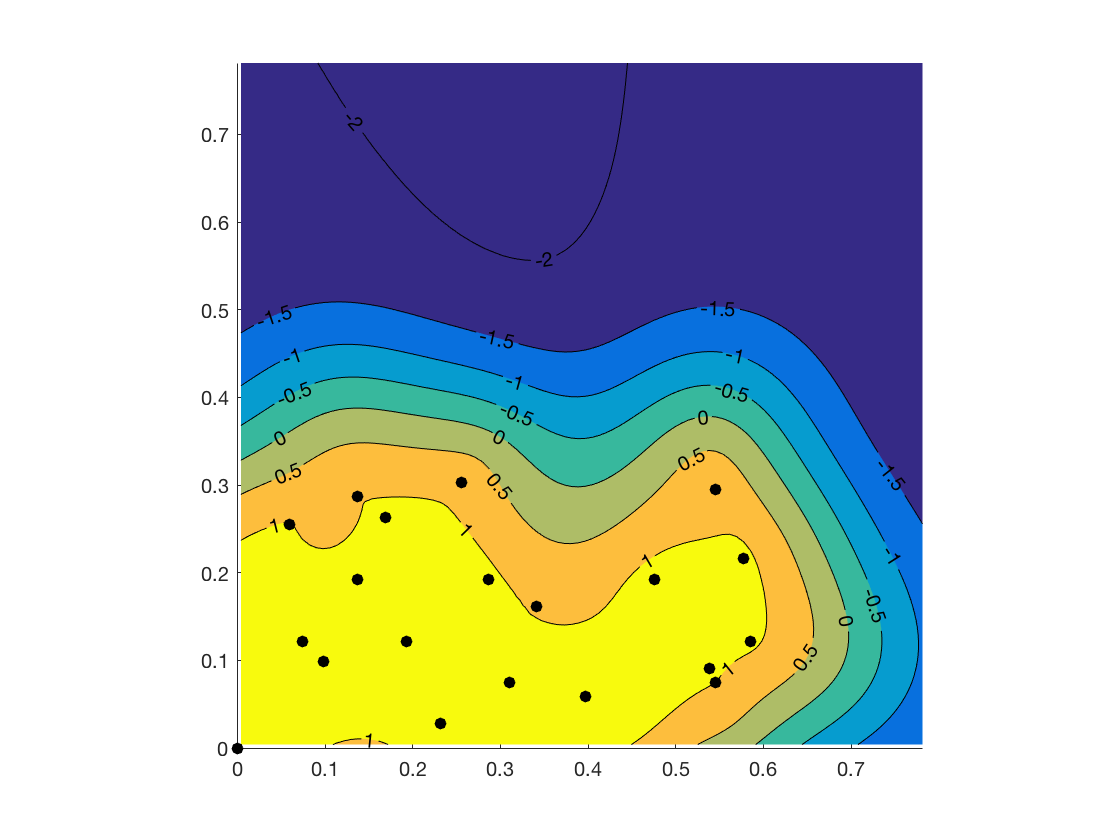}
\includegraphics[width=3.9cm,trim={3cm 0 3cm 0},clip]{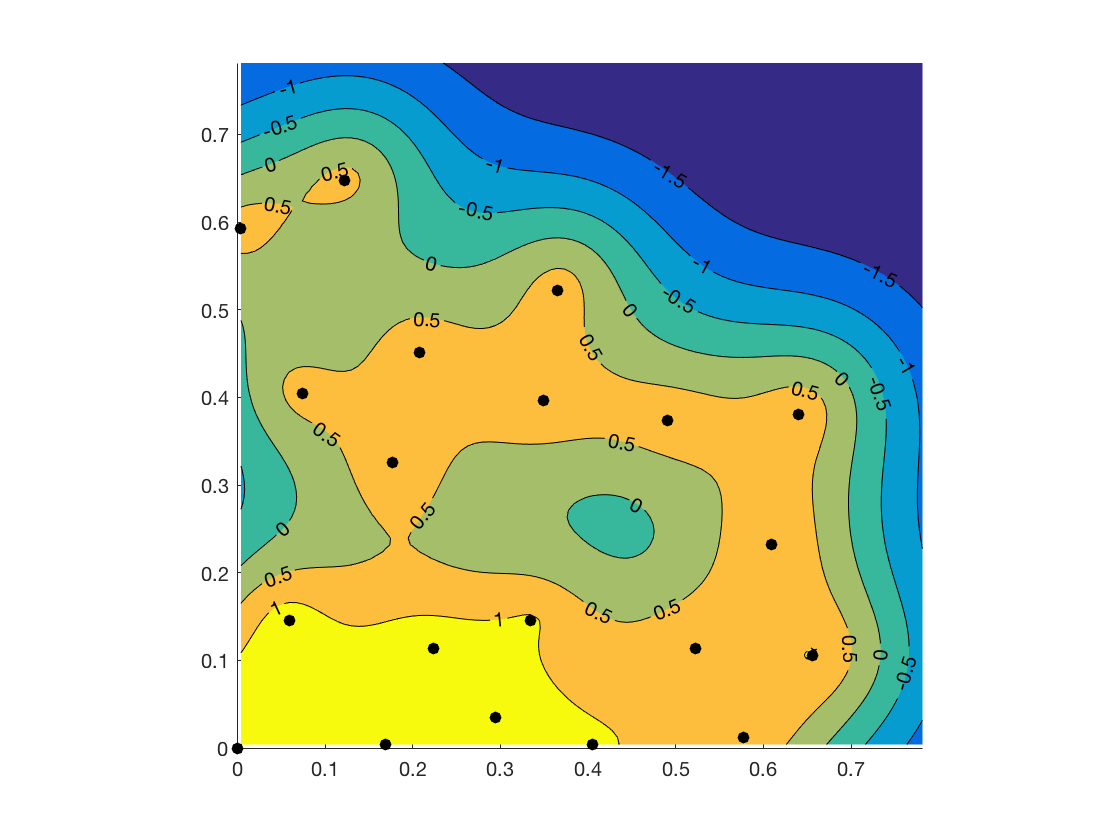}
\includegraphics[width=3.9cm,trim={3cm 0 3cm 0},clip]{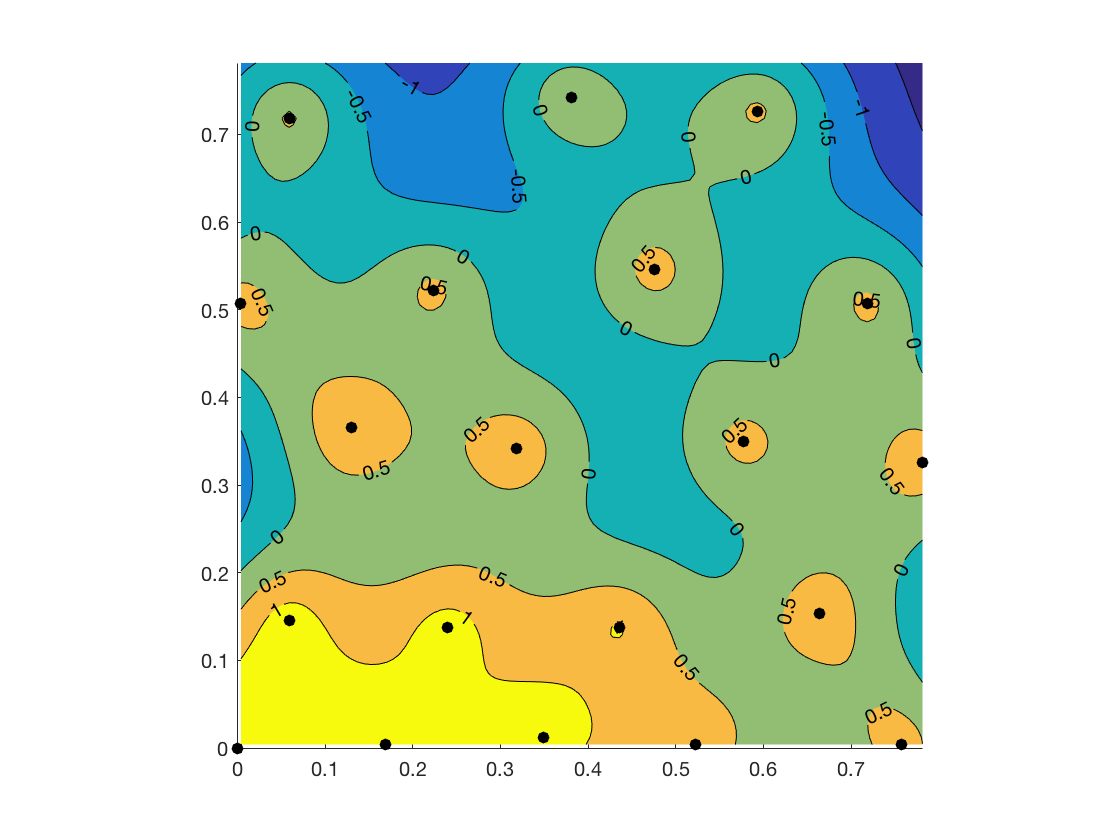}
\caption{Aircraft collision avoidance. Contours for $\mu_{GP}(\mathbf x) - 1.96\sigma_{GP}(\mathbf x)$ given by the posterior Gaussian process. Left: \textsc{RMILE} algorithm. Middle: \textsc{LSE} algorithm. Right: Straddle heuristic. The yellow region is the area of interest.}
\label{fig:Uav}
\end{figure*}

We evaluate our method in the task of estimating the sensor requirements for an aircraft collision avoidance system. We consider pairwise encounters between aircrafts, the behavior of which is dictated by a joint policy produced by modeling the problem as a Markov decision process and solving for optimal actions using value iteration. Observation noise is applied over two state variables, the relative angle and heading between the two aircrafts. 
The noise for each variable is sampled independently from a normal distribution with mean zero and standard deviation varying depending on assumed sensor precision. For each sample, 500 pairwise encounters are simulated, and the estimated probability of a near mid-air collision (NMAC) is returned. We apply the negative logit transformation to the output to map it to the real line. We look for a threshold $t = 1$, and the origin is given as a seed. Again, \textsc{RMILE} samples in a more structured way, progressively expanding $I_{GP}$ while balancing the reduction of the variance in the promising region with some exploration (Figure \ref{fig:Uav}).

\subsection{Simulation experiments: required sensor precision}
We assess our method on estimating actuator performance requirements in an automotive setting. We seek to determine the necessary precision for longitudinal and lateral acceleration maneuvers of simulated vehicles such that the likelihood of hard braking events is below a threshold. In these experiments, we simulate a single, five-second scenario involving twenty vehicles for $100$ steps. The vehicles are propagated according to a bicycle model, with longitudinal behavior generated by the Intelligent Driver Model 
\cite{idm} and lane changing behavior dictated by the MOBIL \cite{mobil} model.  
The two input parameters are sampled from a normal distribution, the standard deviation of which models the actuator precision.
 
\begin{figure*}[h]
\centering
\begin{subfigure}[b]{1\textwidth}
\centering
\includegraphics[width=4.0cm,trim={1.2cm 1cm 1.2cm 1cm},clip]{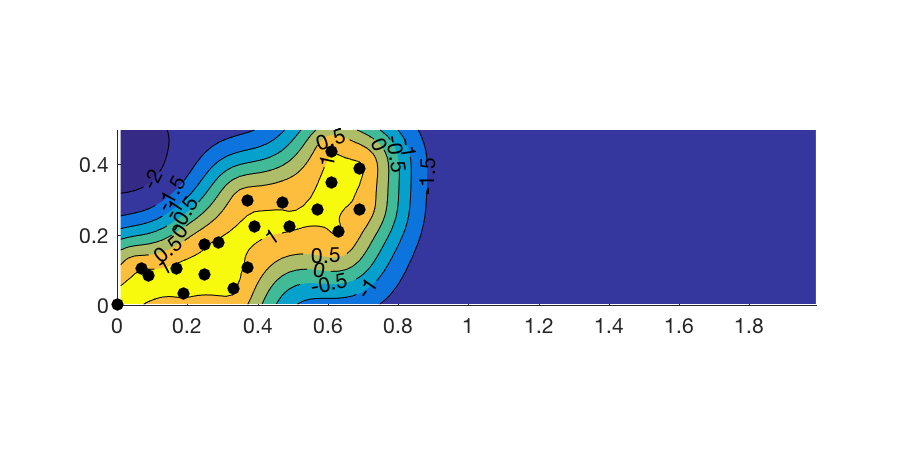}
\includegraphics[width=4.0cm,trim={1.2cm 1cm 1.2cm 1cm},clip]{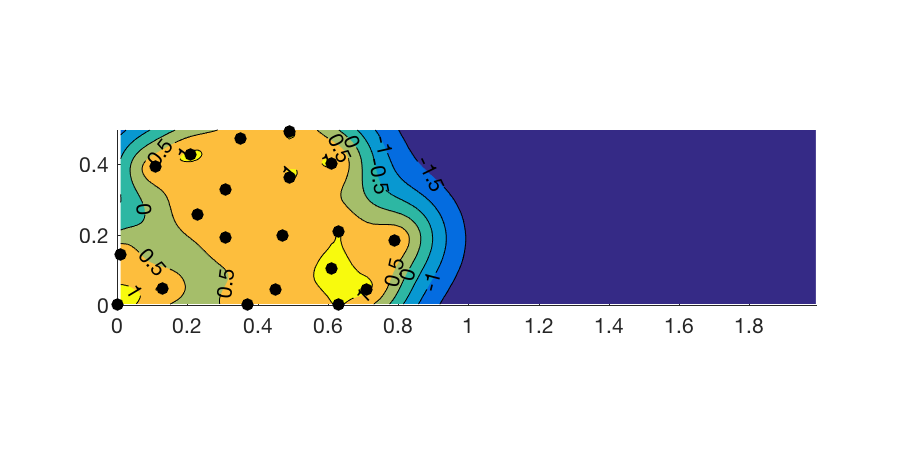}
\includegraphics[width=4.0cm,trim={1.2cm 1cm 1.2cm 1cm},clip]{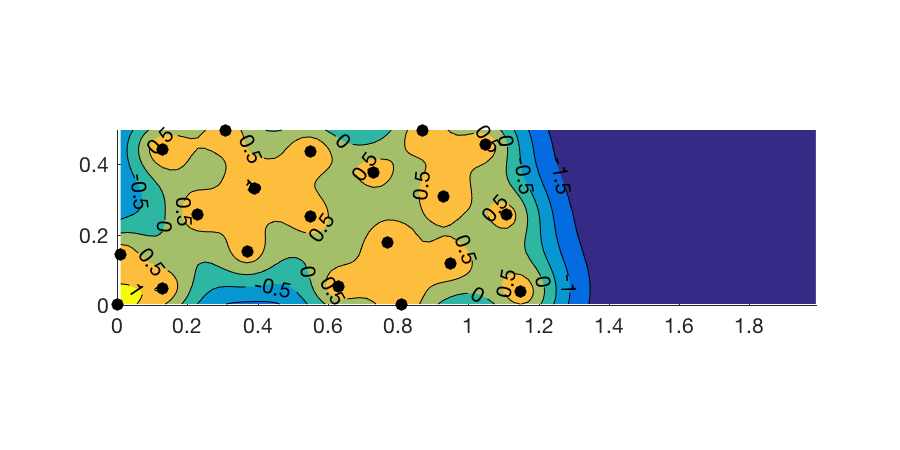}
\caption{$\mu_{GP}(\mathbf x) - 1.96\sigma_{GP}(\mathbf x)$ contours. \textsc{RMILE} on the left, \textsc{LSE} in the middle, and Straddle on the right. The yellow region is the area of interest.}
\label{fig:Auto}		
\end{subfigure}

\begin{subfigure}[b]{1\textwidth}
\centering
\includegraphics[width=4.0cm,trim={1.2cm 1cm 1.2cm 1cm},clip]{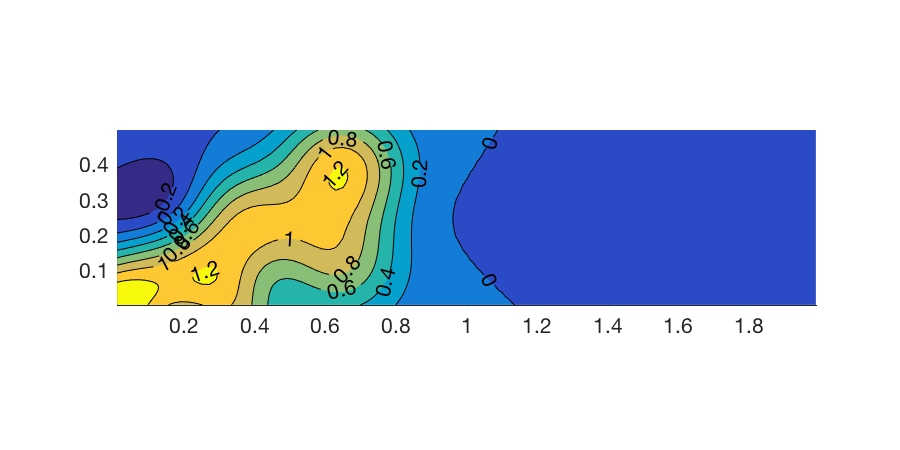}
\includegraphics[width=4.0cm,trim={1.2cm 1cm 1.2cm 1cm},clip]{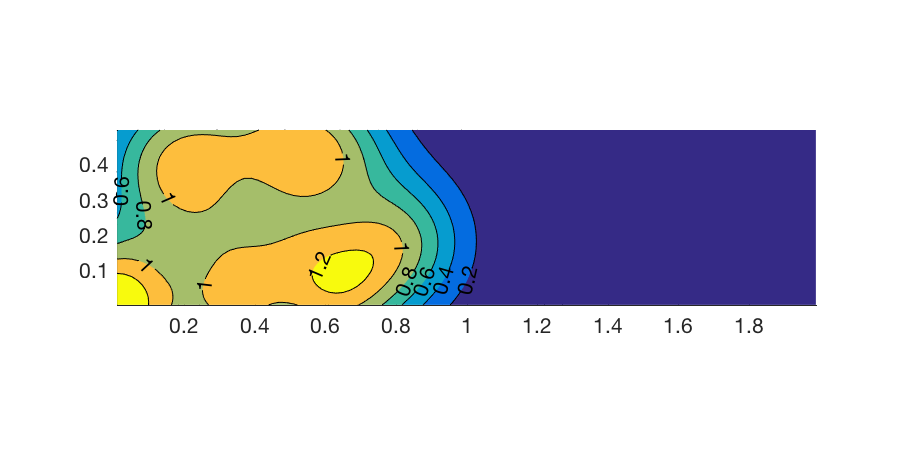}
\includegraphics[width=4.0cm,trim={1.2cm 1cm 1.2cm 1cm},clip]{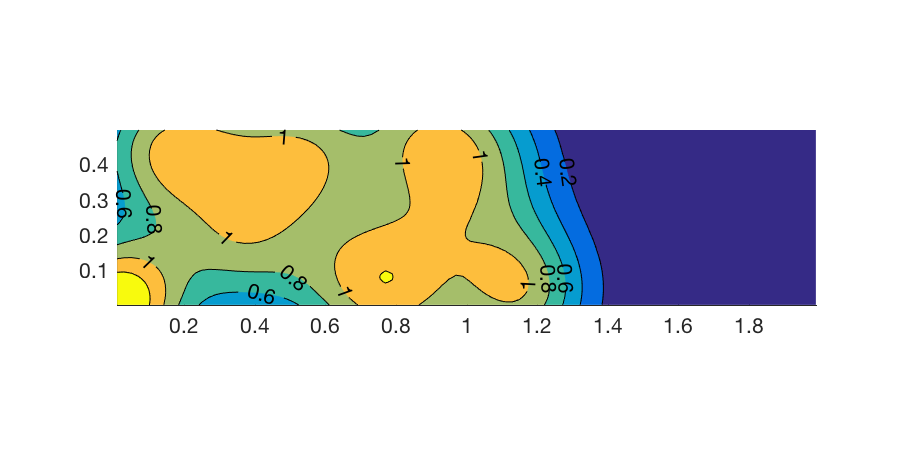}
\caption{$\mu_{GP}(\mathbf x)$ contours. \textsc{RMILE} on the left, \textsc{LSE} in the middle, Straddle on the right.}
\label{fig:Auto_mu}		
\end{subfigure}

\begin{subfigure}[b]{1\textwidth}
\centering
\includegraphics[width=4.0cm,trim={1.2cm 1cm 1.2cm 1cm},clip]{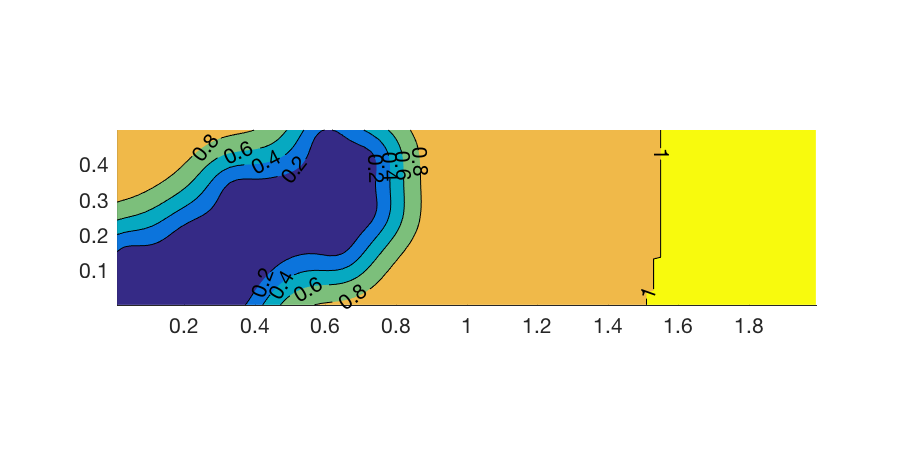}
\includegraphics[width=4.0cm,trim={1.2cm 1cm 1.2cm 1cm},clip]{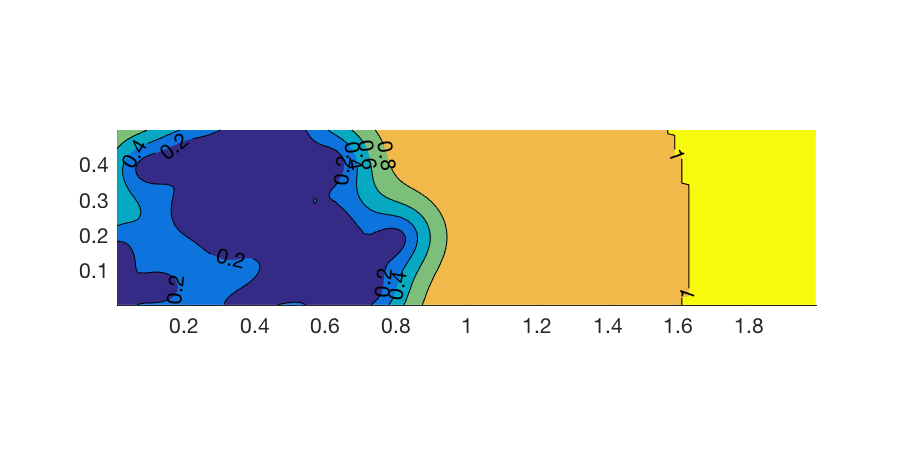}
\includegraphics[width=4.0cm,trim={1.2cm 1cm 1.2cm 1cm},clip]{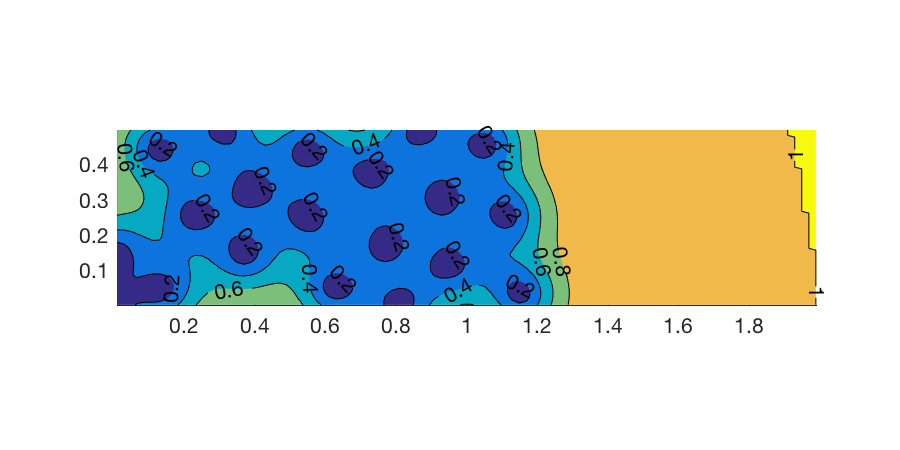}
\caption{$1.96\sigma_{GP}(\mathbf x)$ contours. \textsc{RMILE} on the left, \textsc{LSE} in the middle, Straddle on the right.}
\label{fig:Auto_si}		
\end{subfigure}
\caption{Auto example: required sensor precision.}
\end{figure*}

We model the (estimated) probability of hard braking using the negative logit function $-\log\frac{y}{1-y}$, which maps the outcome $y \in [0,1]$ from the simulator to the real line. This is not strictly needed but ensures that the Gaussian process is consistent with the type of output. We run an exploratory simulation with a budget of 20 points for Straddle, \textsc{LSE} and our algorithm. While the underlying function has some random noises due to the Monte Carlo simulation, we fix the seed to have the same point-wise responses from the simulator when different algorithms are tested. We select a threshold of 1.0 and again choose the origin as the initial seed. In Figure \ref{fig:Auto} we plot the contours for $\mu_{GP}(\mathbf x) - 1.96\sigma_{GP}(\mathbf x)$. It can be seen that \textsc{LSE} and the Straddle heuristic both try to reduce the uncertainty by spreading out the sample points.
Crucially, our algorithm places more samples together to compensate for the noise and reduce the variance (Figure \ref{fig:Auto_si}) in the promising region (Figure \ref{fig:Auto_mu}) above $t = 1.0$ . This allows the classifier to use the posterior GP to make a more confident prediction and identify the area of interest with high confidence (yellow region in Figure \ref{fig:Auto}).

\section{Conclusions}
We have considered the problem of level set estimation where only a noise-corrupted version of the function is available with a very limited exploration budget. The aim is to discover as rapidly as possible a region where the threshold is exceeded with high probability. We propose to select the next query point that maximizes the expected volume of the domain of points above the threshold in a one-step lookahead procedure, and derive analytical formulae to compute this quantity in closed forms. We give a simple criterion to verify convergence of generic acquisition functions and verify that our algorithm satisfies such requirements. Our algorithm also compares favorably with the state of the art on numerical experiments. In particular, it  uses information gained from a few samples more effectively, making it suitable when a very limited exploration budget is available. At the same time, it retains asymptotic convergence guarantees, making it especially compelling in the case of misspecified models. 

\subsubsection*{Acknowledgments}
Blake Wulfe provided the simulator for the simulations experiments. The authors are grateful to the reviewers for their comments.

\medskip
\small

\bibliography{all_bib}

\begin{thebibliography}{10}

\bibitem{Got13}
A.~Gotovos, N.~Casati, G.~Hitz, and A.~Krause.
\newblock Active learning for level set estimation.
\newblock In {\em International Joint Conference on Artificial Intelligence
  (IJCAI)}, pages 1344--1350, 2013.

\bibitem{GPML}
C.~E. Rasmussen and C.~K.~I. Williams.
\newblock {\em Gaussian Processes for Machine Learning}.
\newblock MIT Press, 2006.

\bibitem{Fre16}
B.~Shahriari, K.~Swersky, Z.~Wang, R.~P. Adams, and N.~de Freitas.
\newblock Taking the human out of the loop: A review of {B}ayesian
  optimization.
\newblock 104(1):148 -- 175, 2016.

\bibitem{Kush64}
H.~J. Kushner.
\newblock A new method of locating the maximum point of an arbitrary multipeak
  curve in the presence of noise.
\newblock {\em Journal of Basic Engineering}, 86(1), 1964.

\bibitem{Mock78}
J.~Mockus, V.~Tiesis, and A.~Zilinskas.
\newblock The application of {B}ayesian methods for seeking the extremum.
\newblock In L.~C.~W. Dixon and G.~P. Szego, editors, {\em Towards Global
  Optimization}, volume~2. North-Holland Publishing Company, 1978.

\bibitem{Sri10}
N.~Srinivas, A.~Krause, S.~M. Kakade, and M.~Seeger.
\newblock Gaussian process optimization in the bandit setting: No regret and
  experimental design.
\newblock In {\em International Conference on Machine Learning (ICML)}, pages
  1015--1022, 2010.

\bibitem{shah14}
B.~Shahriari, Z.~Wang, M.~W. Hoffman, A.~Bouchard-Cote, and N.~de Freitas.
\newblock An entropy search portfolio for {B}ayesian optimization.
\newblock In {\em Advances on Neural Information Processing Systems (NIPS)},
  2014.

\bibitem{Wil07}
R.~M. Willet and R.~D. Nowak.
\newblock Minimax optimal level-set estimation.
\newblock {\em IEEE Transaction on Image Processing}, 16(12):2965--2979, 2007.

\bibitem{Son12}
A.~Soni and J.~Haupt.
\newblock Level set estimation from compressive measurements using box
  constrained total variation regularization.
\newblock In {\em IEEE International Conference on Image Processing (ICIP)},
  pages 2573--2576, 2012.

\bibitem{Kri13}
K.~Krishnamurthy, W.~U. Bajwa, and R.~Willett.
\newblock Level set estimation from projection measurements: Performance
  guarantees and fast computation.
\newblock {\em SIAM Journal on Imaging Sciences}, 6(4):2047--2074, 2013.

\bibitem{Kra08}
A.~Krause, A.~Singh, and C.~Guestrin.
\newblock Near-optimal sensor placements in {G}aussian processes: Theory,
  efficient algorithms and empirical studies.
\newblock {\em Journal of Machine Learning Research}, 9:235--284, 2008.

\bibitem{Mar17}
L.~Martino, J.~Vicent, and G.~Camps-Valls.
\newblock Automatic emulator and optimized look-up table generation for
  radiative transfer models.
\newblock In {\em IEEE International Geoscience and Remote Sensing Symposium},
  pages 1457--1460, 2017.

\bibitem{Bu09}
D.~Busby.
\newblock Hierarchical adaptive experimental design for {G}aussian process
  emulators.
\newblock {\em Reliability Engineering \& System Safety}, 94(7):1183 -- 1193,
  2009.

\bibitem{Bry05}
B.~Bryan, R.~C. Nichol, C.~R. Genovese, J.~Schneider, C.~J. Miller, and
  L.~Wasserman.
\newblock Active learning for identifying function threshold boundaries.
\newblock In {\em Advances in neural information processing systems (NIPS)},
  pages 163--170, 2005.

\bibitem{Bog16}
I.~Bogunovic, J.~Scarlett, A.~Krause, and V.~Cevher.
\newblock Truncated variance reduction: A unified approach to {B}ayesian
  optimization and level-set estimation.
\newblock In {\em Advances in neural information processing systems (NIPS)},
  pages 1507--1515, 2016.

\bibitem{Yang14}
J.~Yang, Z.~Wang, and J.~Wu.
\newblock Level set estimation with dynamic sparse sensing.
\newblock In {\em IEEE Global Conference on Signal and Information Processing
  (GlobalSIP)}, pages 487--491, 2014.

\bibitem{Ma14}
Y.~Ma, R.~Garnett, and J.~Schneider.
\newblock Active area search via {B}ayesian quadrature.
\newblock In {\em Artificial Intelligence and Statistics (AISTATS)}, pages
  595--603, 2014.

\bibitem{Ma15}
Y.~Ma, D.~Sutherland, R.~Garnett, and J.~Schneider.
\newblock Active pointillistic pattern search.
\newblock In {\em Artificial Intelligence and Statistics (AISTATS)}, pages 672
  -- 680, 2015.

\bibitem{Bect12}
J.~Bect, D.~Ginsbourger, L.~Li, V.~Picheny, and E.~Vazquez.
\newblock Sequential design of computer experiments for the estimation of a
  probability of failure.
\newblock {\em Statistics and Computing}, 22(3):773--793, 2012.

\bibitem{Cheval14}
C.~Chevalier, J.~Bect, D.~Ginsbourger, E.~Vazquez, V.~Picheny, and Y.~Richet.
\newblock Fast parallel kriging-based stepwise uncertainty reduction with
  application to the identification of an excursion set.
\newblock {\em Technometrics}, 56(4):455--465, 2014.

\bibitem{idm}
M.~Treiber, A.~Hennecke, and D.~Helbing.
\newblock Congested traffic states in empirical observations and microscopic
  simulations.
\newblock 62(2):1805, 2000.

\bibitem{mobil}
A.~Kesting, M.~Treiber, and D.~Helbing.
\newblock General lane-changing model {MOBIL} for car-following models.
\newblock {\em Transportation Research Record}, 1999(1):86--94, 2007.

\end{thebibliography}
\bibliographystyle{unsrt}

\newpage
\section*{Appendix A: Technical Proofs}

\textbf{Lemma \ref{lmm:classification}.}
The classification rule identified by  Eqn. (\ref{classificationrulebis}) minimizes the expected weighted misclassification error:
\begin{equation}
\E \left( \delta \mathbbm{1}\{\mathbf x \in \text{FP}\} + (1-\delta) \mathbbm{1}\{\mathbf x \in \text{FN}\} \right),
\end{equation}
among all deterministic classification rules under the posterior probability measure given by the Gaussian processes. Here $\mathbf x\in \Omega$ is arbitrary and fixed.\\

\noindent\textbf{Proof:} To prove the optimality of our classification rule (2) and (3), consider an arbitrary fixed point $x\in \Omega$. We proceed by a case-by-case analysis.

\textbf{Case I:} suppose that $P_{GP}(f(\mathbf x)>t)>\delta$ is satisfied. Then if $\mathbf x$ is classified into $I_{GP}$, we have
\begin{align}
\E &\left( \delta \mathbbm{1}\{\mathbf x \in \text{FP}\} + (1-\delta) \mathbbm{1}\{\mathbf x \in \text{FN}\} \right) \nonumber\\
&=\delta P(\mathbf x\in\text{FP})+(1-\delta) P(\mathbf x\in\text{FN})\\
&=\delta P_{GP}(f(\mathbf x)\leq t) <\delta(1-\delta) \nonumber
\end{align}
and if $\mathbf x$ is not classified into $I_{GP}$, then we have
\begin{align}
\E &\left( \delta \mathbbm{1}\{\mathbf x \in \text{FP}\} + (1-\delta) \mathbbm{1}\{\mathbf x \in \text{FN}\} \right)\\
&=(1-\delta) P_{GP}(f(\mathbf x)> t) >(1-\delta)\delta \nonumber
\end{align}
which implies that classifying $\mathbf x$ into $I_{GP}$ is better than not, given that $P_{GP}(f(\mathbf x)>t)>\delta$.

\textbf{Case II:} suppose that $P_{GP}(f(\mathbf x)>t)\leq\delta$. Then if $\mathbf x$ is classified into $I_{GP}$, we have
\begin{align}
\E &\left( \delta \mathbbm{1}\{\mathbf x \in \text{FP}\} + (1-\delta) \mathbbm{1}\{\mathbf x \in \text{FN}\} \right)\\
&=\delta P_{GP}(f(\mathbf x)\leq t) \geq\delta(1-\delta) \nonumber
\end{align}
and if $\mathbf x$ is not classified into $I_{GP}$, then we have
\begin{align}
\E &\left( \delta \mathbbm{1}\{\mathbf x \in \text{FP}\} + (1-\delta) \mathbbm{1}\{\mathbf x \in \text{FN}\} \right)\\
&=(1-\delta) P_{GP}(f(\mathbf x)> t) \leq (1-\delta)\delta \nonumber
\end{align}
which implies that classifying $\mathbf x$ into $I_{GP}$ is worse than not, given that $P_{GP}(f(\mathbf x)>t)\leq\delta$.

Hence, we conclude that our classification rule minimizes the expected weighted misclassification error given in Lemma 1 for any fixed point $\mathbf x\in \Omega$.
\qed \\

\noindent\textbf{Lemma \ref{fast_compute}.}
$ \E_{y^+} \left|I_{GP^+}\right|$ obtained by sampling at $\mathbf x^+$ can be computed analytically as follows:
\begin{equation}
\sum_{\mathbf x \in \Omega} \Phi \left(\frac{\sqrt{\sigma^2_{GP}(\mathbf x^+) + \sigma^2_{\epsilon}
 }}{|\textup{Cov}_{GP}(f(\mathbf x), f(\mathbf x^+))|} \times \left( \mu_{GP}(\mathbf x) - \beta\sigma_{GP^+}(\mathbf x) -t  \right) \right) 
\label{objective}
\end{equation}
where $\Phi(\cdot)$ is the cumulative distribution function (CDF) of the standard normal random variables, and 
\begin{equation}
\sigma^2_{GP^+}(\mathbf x)=\textup{Cov}_{GP^+}(f(\mathbf x),f(\mathbf x)) = \sigma^2_{GP}(\mathbf x) - \dfrac{\textup{Cov}_{GP}^2(f(\mathbf x), f(\mathbf x^+))}{\sigma^2_{GP}(\mathbf x^+) + \sigma^2_{\epsilon}},
\end{equation}
and $\textup{Cov}_{GP}(f(\mathbf x), f(\mathbf x^+))=k_{GP}(\mathbf x,\mathbf x^+)$ is the (current) posterior covariance between $f(\mathbf x)$ and $f(\mathbf x^+)$.\\

\noindent\textbf{Proof:}
Let $f_{GP}(y^+)$ be marginal density of $y^+$, which is a normal PDF. Then by definition of $\left|I_{GP^+}\right|$, we have that
\begin{align}
\E_{y^+}\left|I_{GP^+}\right| = & \E_{y^+} \sum_{\mathbf x \in \Omega} \mathbbm{1} \left\{ \mu_{GP^+} (\mathbf x ; y^+) - \beta \sigma_{GP^+} (\mathbf x ) > t \right\} \nonumber\\
= & \sum_{\mathbf x \in \Omega} \E_{y^+} \mathbbm{1} \left\{ \mu_{GP^+}  (\mathbf x ; y^+) - \beta \sigma_{GP^+} (\mathbf x ) > t \right\}\nonumber \\
= & \sum_{\mathbf x \in \Omega} P_{GP}\left(\mu_{GP^+} (\mathbf x ; y^+) - \beta \sigma_{GP^+} (\mathbf x) > t \right) \nonumber\\
= & \sum_{\mathbf x \in \Omega} \int_{-\infty}^{+\infty} P_{GP}(\mu_{GP^+} (\mathbf x ; y^+) - \beta \sigma_{GP^+} (\mathbf x ) > t | y^+) f_{GP}(y^+) dy^+  \nonumber\\
= & \sum_{\mathbf x \in \Omega} \int_{-\infty}^{+\infty} \mathbbm{1} \{ \mu_{GP^+} (\mathbf x ; y^+) \nonumber - \beta \sigma_{GP^+} (\mathbf x ) > t | y^+ \} f_{GP}(y^+) dy^+ \nonumber
\end{align}
where the last equality immediately follows from the fact that conditioned on $y^+$, the event $\left\{ \mu_{GP^+} (\mathbf x ; y^+) - \beta \sigma_{GP^+} (\mathbf x ; y^+) > t \right \}$ is either true or false and thus the probability can be recast as an indicator.

Furthermore, we have that  any two random variable $f(\mathbf x^+), f(\mathbf x)$ identified by a Gaussian process follow a bivariate normal distribution and the posterior variance is independent from $y^+$ at $\mathbf x$ (\textit{cf}. C.E. Russmussen, 2006, or simply Eqns. (\ref{eqn:GP})):
\begin{align}
\sigma^2_{GP^+}(\mathbf x) = \sigma^2_{GP}(\mathbf x) - \frac{\textup{Cov}_{GP}^2(f(\mathbf x), f(\mathbf x^+))}{\sigma^2_{GP}(\mathbf x^+)+\sigma^2_{\epsilon}}
\end{align}
The posterior mean can be computed as
\begin{align}\label{mu-gp+}
\mu_{GP^+}(\mathbf x; y^+) = \mu_{GP}(\mathbf x) - \frac{\textup{Cov}_{GP}(f(\mathbf x), f(\mathbf x^+))}{\sigma^2_{GP}(\mathbf x^+)+\sigma^2_{\epsilon}}(y^+ - \mu_{GP}(\mathbf x^+))
\end{align}

Let $y^L_{GP}(\mathbf{x},\mathbf x^+) $ be the limit value for the outcome $y^+$ at $\mathbf x$ such that $\mu_{GP^+} (\mathbf x) - \beta \sigma_{GP^+} (\mathbf x) = t $. Plugging in the expression for the posterior variance and mean yields:
\begin{equation}
y^L_{GP}(\mathbf{x},\mathbf x^+ ) = \frac{\sigma^2_{GP}(\mathbf x^+)+\sigma^2_{\epsilon}}{\textup{Cov}_{GP}(f(\mathbf x),f(\mathbf x^+))}
\times\left(t + \beta\sigma_{GP^+}(\mathbf x) - \mu_{GP}(\mathbf x)\right) +  \mu_{GP}(\mathbf x^+) 
\label{eqnapp:yLim}
\end{equation}

Thus the objective function $\E_{y^+}\left|I_{GP^+}\right|$ can be recast as:

\begin{equation}
\begin{split}
&\sum_{\mathbf x \in \Omega : \textup{Cov}_{GP}(f(\mathbf x), f(\mathbf x^+)) \geq 0} \int_{y^L_{GP}(\mathbf{x},\mathbf x^+ ) }^{+\infty} f_{GP}(y^+) dy^+ \\
&+ \sum_{\mathbf x \in \Omega : \textup{Cov}_{GP}(f(\mathbf x), f(\mathbf x^+)) < 0} \int_{-\infty}^{y^L_{GP}(\mathbf{x},\mathbf x^+ ) } f_{GP}(y^+) dy^+
\end{split}
\end{equation}
where summations are over terms with positive, respectively negative covariances. Finally we can combine the case with positive and negative covariances into one and move from the normal density $f_{GP}$ to a standard normal CDF by subtracting the mean $\mu_{GP}(\mathbf x^+)$ and dividing by the marginal variance of $y^+$  (which is $\sqrt{\sigma^2_{GP}(\mathbf x^+) + \sigma^2_{\epsilon}}$) to obtain:
\begin{equation}
\sum_{\mathbf x \in \Omega} \Phi \left(\frac{\sqrt{\sigma^2_{GP}(\mathbf x^+) + \sigma^2_{\epsilon}
 }}{|\textup{Cov}_{GP}(f(\mathbf x), f(\mathbf x^+))|} \times \left( \mu_{GP}(\mathbf x) - \beta\sigma_{GP^+}(\mathbf x) -t  \right) \right) 
\end{equation}
\qed \\

\noindent\textbf{Lemma \ref{lemma:varmin}.}
If the acquisition function is redefined as:
\begin{equation}\label{eqn:varred}
\begin{split}
E_{GP}^{var}(\mathbf x^+):=&\E_{y^+}\int_{-\infty}^{\infty} |I^{(t)}_{GP^+}| - |I^{(t)}_{GP}| dt,
\end{split}
\end{equation}
then Algorithm 2 minimizes the $l_1$-norm of the posterior standard deviation, \textit{i.e.}, the next query point $\mathbf x^+$ is selected as $\mathbf x^+=\argmin_{\mathbf x^+ \in \Omega} \sum_{\mathbf x \in \Omega} \sigma_{GP^+}(\mathbf x)$.\\

\noindent\textbf{Proof:} By definition, we have that
\begin{equation*}
\begin{split}
\E_{y^+} &\sum_{\mathbf x \in \Omega} \int_{-\infty}^{\infty} \mathbbm{1} \left\{ \mu_{GP^+} (\mathbf x;y^+) - \beta \sigma_{GP^+} (\mathbf x) > t \right\} - \mathbbm{1} \left\{ \mu_{GP} (\mathbf x) - \beta \sigma_{GP} (\mathbf x) > t \right\} dt \\
&= \E_{y^+} \sum_{\mathbf x \in \Omega} \int_{\mu_{GP} (\mathbf x) - \beta \sigma_{GP}(\mathbf x)}^{\mu_{GP^+} (\mathbf x;y^+) - \beta \sigma_{GP^+}(\mathbf x)}  dt\\
&= \E_{y^+} \sum_{\mathbf x \in \Omega} \mu_{GP^+} (\mathbf x;y^+) - \beta \sigma_{GP^+}(\mathbf x) -
\mu_{GP} (\mathbf x) + \beta \sigma_{GP}(\mathbf x) \\
&=   \sum_{\mathbf x \in \Omega} \E_{y^+} \left( \mu_{GP^+} (\mathbf x) - \mu_{GP} (\mathbf x) \right ) + \beta \sigma_{GP}(\mathbf x) -
 \beta \sigma_{GP^+}(\mathbf x)\\
&= -\beta \sum_{\mathbf x \in \Omega} \left(\sigma_{GP^+}(\mathbf x) -  \sigma_{GP}(\mathbf x) \right),
\end{split}
\end{equation*}
where the first passage follows from the fact that the integrand is $1$ or $-1$ only when $ \mu_{GP^+} (\mathbf x;y^+) - \beta \sigma_{GP^+} (\mathbf x) > t > \mu_{GP} (\mathbf x) - \beta \sigma_{GP} (\mathbf x) $ or when  $ \mu_{GP^+} (\mathbf x;y^+) - \beta \sigma_{GP^+} (\mathbf x) < t < \mu_{GP} (\mathbf x) - \beta \sigma_{GP} (\mathbf x) $, respectively.
The last passage follows from $\mu_{GP} (\mathbf x;y^+) = \E_{y^+} \mu_{GP^+} (\mathbf x)$ by linearity (\textit{cf}. Eqn. (\ref{mu-gp+})) and $\E_{y^+} \sigma_{GP^+} (\mathbf x) = \sigma_{GP^+} (\mathbf x)$ because the outcome $y^+$ does not affect the posterior variance, once the sampling location $\mathbf x^+$ is fixed.
Thus 
\begin{equation}
\argmax_{\mathbf x^+ \in \Omega} \E_{y^+}\int_{-\infty}^{\infty} |I^{(t)}_{GP^+}| - |I^{(t)}_{GP}| dt  
\end{equation}
 is equivalent to
\begin{equation}
\argmin_{\mathbf x^+ \in \Omega} \sum_{\mathbf x \in \Omega} \sigma_{GP^+} (\mathbf x) 
\end{equation}
or more explicitly
\begin{equation}
\argmin_{\mathbf x^+ \in \Omega} \sum_{\mathbf x \in \Omega} \sqrt{\sigma^2_{GP}(\mathbf x) - \frac{\textup{Cov}_{GP}^2(f(\mathbf x), f(\mathbf x^+))}{\sigma^2_{GP}(\mathbf x^+) + \sigma^2_{\epsilon}}}
\end{equation}
\qed\\

\noindent\textbf{Lemma \ref{lemma:convergence}.}
Let $E_{GP}(\mathbf x^+)$ be an acquisition function that depends on the posterior $GP$ at a potential query point $\mathbf x^+$, such that for sufficiently small $\sigma^2_{GP}(\mathbf x^+)$, there exists a function $u(\cdot)$ which only depends on the posterior variance $\sigma^2_{GP}(\mathbf x^+)$, with
\begin{equation}
\label{eqn:bracket}
E_{GP}(\mathbf x^+) \leq u(\sigma_{GP}(\mathbf x^+)),\quad \lim_{\sigma(\mathbf x^+)\to 0^+} u(\sigma_{GP}(\mathbf x^+)) = 0.
\end{equation}
In addition, assume that there exists a global lower bound $l(\cdot)$, such that 
\begin{equation} 
E_{GP}(\mathbf x^+)\geq l(\sigma_{GP}(\mathbf x^+)),\quad \lim_{\sigma(\mathbf x^+)\to 0^+} l(\sigma_{GP}(\mathbf x^+)) = 0,
\end{equation}
and assume that $l(\sigma_{GP}(\mathbf x^+))$ is strictly increasing in $\sigma_{GP}(\mathbf x^+)$.

If Algorithm \ref{alg:Framwork} selects the next query point as $\argmax_{\mathbf x^+} E_{GP}(\mathbf x^+) $ and is run without termination, then there cannot be a point in the grid that is sampled only finitely many times.\\

{\noindent\textbf{Proof:} In this proof, we go back to the notation in Section 2.1 in order to keep track of the entire sequence of observations, which facilitates the discussions about limiting behavior here.

We begin by summarizing some basic properties about Gaussian processes (GP) and the multivariate normal distributions. 
\begin{itemize}
\item By definition, a GP with (prior) mean $\mu_0({\mathbf x})$ and kernel $k_0({\mathbf x},{\mathbf x'})$ identifies a multivariate normal distribution on the fixed finite grid ${\mathbf z_1},\dots, {\mathbf z_m}$ with 
\[
\boldsymbol{\mu}^0=[\mu_0({\mathbf z_1}),\dots,\mu_0({\mathbf z_m})]^T,\quad {\mathbf \Sigma}^0=[k_0({\mathbf z_i},{\mathbf z_j}))]_{i,j=1,\dots, m},
\] as the mean vector and the covariance matrix, respectively.
\item The posterior GP  after observing $n$ noisy observations ${\mathbf x_1},\dots, {\mathbf x_n}$ (with possible repetitions) again identifies a multivariate normal distribution on the fixed finite grid ${\mathbf z_1},\dots, {\mathbf z_m}$ with 
\[
\boldsymbol{\mu}^n=[\mu_n({\mathbf z_1}),\dots,\mu_0({\mathbf z_m})]^T,\quad {\mathbf \Sigma}^n=[k_n({\mathbf z_i},{\mathbf z_j}))]_{i,j=1,\dots, m},
\] as the mean vector and the covariance matrix, respectively. 
\item (\emph{Exchangeability}) Moreover, the order of observations does not affect the posterior distribution. More explicitly, we have for all $i,j=1,\dots,m$,
\[
\mu_{\mathbf x_{1:n},y_{1:n}}({\mathbf z}_i)=\mu_{\mathbf x_{\pi(1:n)},y_{\pi(1:n)}}({\mathbf z}_i),\quad k_{\mathbf x_{1:n}}({\mathbf z_i},{\mathbf z_j})=k_{\mathbf x_{\pi(1:n)}}({\mathbf z}_i,{\mathbf z}_j),
\]
where $\pi(1:n)$ is an arbitrary permutation of $1:n$. 
\item (\emph{Associativity}) In addition, we also have the following associativity property: if we replace $\mu_0$ and $k_0$ with $\mu_{\mathbf{x}_{1:n_1}}$ and $k_{\mathbf{x}_{1:n_1}}$, and apply Eqns. (\ref{eqn:GP}) to $n-n_1$ new samples $\mathbf{x}_{n_1+1},\dots,\mathbf{x}_{n}$, then we obtain exactly $\mu_n$ and $k_n$.
\item Starting from any prior kernel $k_0$, if a point $\bar{\mathbf x}\in \{\mathbf{z}_{1:m}\}$ is sampled $n$ times (\textit{i.e.}, $\mathbf{x}_1=\dots=\mathbf{x}_n=\bar{\mathbf x}$), then the posterior kernel/covariance in Eqns. (\ref{eqn:GP}) is equal to 
\begin{equation}\label{post_formula}
k_n(\mathbf{z}_i,\mathbf{z}_j)=k_0(\mathbf{z}_i,\mathbf{z}_j)-\frac{k_0(\mathbf z_i,\bar{\mathbf{x}})k_0(\bar{\mathbf{x}},\mathbf{z}_j)}{k_0(\bar{\mathbf{x}},\bar{\mathbf x})+\sigma_{\epsilon}^2/n}, \quad \forall i,j=1,\dots,m.
\end{equation}
In paritcular, we have
\begin{equation}\label{var_update}
\sigma_n^2(\mathbf{z}_i)=\sigma_n^2(\mathbf{z}_i)-\frac{k_0(\mathbf{z}_i,\bar{\mathbf{x}})^2}{\sigma_0^2(\bar{\mathbf{x}})+\sigma_{\epsilon}^2/n}, \quad \forall i=1,\dots,m.
\end{equation}
\end{itemize}
The first claim is by definition. To see the other four claims, the key observation is that the prior joint distribution of the noise-less function values $f(\mathbf{z}_{1:m})$ at the grid points, as well as the noisy observations at locations $\mathbf x_{1:n}$ (with possible repetitions), is in fact 
\begin{equation*}
\mathcal{N}(\mu_0([\mathbf{z}_{1:m};\mathbf{x}_{1:n}]),k_0([\mathbf{z}_{1:m};\mathbf{x}_{1:n}], [\mathbf{z}_{1:m};\mathbf{x}_{1:n}])+\sigma_{\epsilon}^2\textbf{diag}(\mathbf{0},I_n)),
\end{equation*}
which comes from the independence of the noises immediately. Applying the well-known formula for computing the conditional mean and covariance of multivariate normal distributions (the Schur complement formula), we immediately see that $\boldsymbol{\mu}^n$ and $\mathbf{\Sigma}^n$ above, as computed by Eqns. (\ref{eqn:GP}), characterize exactly the posterior distribution of the GP at $\mathbf{z}_{1:m}$ conditioned on noisy observations at $\mathbf{x}_{1:n}$.  Exchangeability and associativity then follow as natural results of the corresponding properties of conditioning. To see the last property, simply notice that exchangeability and associativity imply that $k_n(\mathbf{x},\mathbf{x}')$ can be alternatively computed by applying Eqns. (\ref{eqn:GP}) with $n=1$ recursively on observations/measurements at the same location $\bar{\mathbf{x}}$ for $n$ times. A simple proof by induction shows the desired formula (\ref{post_formula}).

More explicitly, the case when $n=1$ in (\ref{post_formula}) is a trivial application of Eqns. (\ref{eqn:GP}). Now suppose that the case when $n=l$ is proved. For $n=l+1$, by associativity, we can apply Eqns. (\ref{eqn:GP}) once on $k_l$, yielding
\begin{equation*}
\begin{split}
k_{l+1}(\mathbf{z}_i,\mathbf{z}_j)&=k_l(\mathbf{z}_i,\mathbf{z}_j)-\frac{k_l(\mathbf{z}_i,\bar{\mathbf{x}})k_l(\bar{x},\mathbf{z}_j)}{k_l(\bar{\mathbf{x}},\bar{\mathbf{x}})+\sigma^2_{\epsilon}}.
\end{split}
\end{equation*}
Plugging in 
\begin{equation*}
\begin{split}
&k_l(\mathbf{z}_i,\mathbf{z}_j)=k_0(\mathbf{z}_i,\mathbf{z}_j)-\frac{k_0(\mathbf z_i,\bar{\mathbf{x}})k_0(\bar{\mathbf{x}},\mathbf{z}_j)}{k_0(\bar{\mathbf{x}},\bar{\mathbf x})+\sigma_{\epsilon}^2/l},\quad k_l(\bar{\mathbf{x}},\bar{\mathbf{x}})=\frac{k_0(\bar{\mathbf{x}},\bar{\mathbf{x}})\sigma_{\epsilon}^2/l}{k_0(\bar{\mathbf{x}},\bar{\mathbf x})+\sigma_{\epsilon}^2/l},\\
&k_l(\mathbf{z}_i,\bar{\mathbf{x}})=\frac{k_0(\mathbf{z}_i,\bar{\mathbf{x}})\sigma_{\epsilon}^2/l}{k_0(\bar{\mathbf{x}},\bar{\mathbf x})+\sigma_{\epsilon}^2/l},\quad k_l(\bar{\mathbf{x}},\mathbf{z}_j)=\frac{k_0(\bar{\mathbf{x}},\mathbf{z}_j)\sigma_{\epsilon}^2/l}{k_0(\bar{\mathbf{x}},\bar{\mathbf x})+\sigma_{\epsilon}^2/l},\\
\end{split}
\end{equation*}
we immediately arrive at equation (\ref{post_formula}). And hence the proof is done by induction.

Now we are ready to establish upper and lower bounds on the posterior variance $\sigma_{GP}^2({\mathbf x}^+)=k_n({\mathbf x}^+,{\mathbf x}^+)=\sigma_n^2({\mathbf x}^+)$ for any $\mathbf x^+\in \{{\mathbf z}_1,\dots,{\mathbf z}_m\}$. More explicitly, we will show that if ${\mathbf x}^+$ is sampled $K$ times, then $\sigma_{GP}({\mathbf x}^+)=\Omega(1/K)$ as $K\rightarrow\infty$.  

In fact, consider an arbitrary fixed grid-point, which we denote w.l.o.g. as $\mathbf z_m$. Suppose that $\mathbf z_m$ is sampled $K$ times until step $n$. By exchangeability and associativity, we can assume w.l.o.g. that the other $n-K$ sampling takes place first, yielding a posterior variance $\sigma_{n-K}^2(\mathbf z_m)$, and then we sample $\mathbf z_m$ for $K$ consecutive times starting from $\sigma_{n-K}^2(\mathbf z_m)$ as a prior variance. This means that we have by (\ref{var_update}) that
\begin{equation}\label{decomp}
\sigma_n^2(\mathbf z_m)=\sigma_{n-K}^2(\mathbf{z}_m)-\frac{\sigma_{n-K}^4(\mathbf{z}_m)}{\sigma_{n-K}^2(\mathbf{z}_m)+\sigma_\epsilon^2/K}=\frac{\sigma_{\epsilon}^2}{K+\sigma_{\epsilon}^2/\sigma_{n-K}^2(\mathbf z_m)}.
\end{equation}

Again by (\ref{var_update}), we know that the posterior variance at $\mathbf z_m$ monotonically decreases whenever we sample at some grid point. Hence we see that $\sigma_{n-K}^2(\mathbf z_m)$ is no larger than when all other grid points are not sampled, which would yield a posterior variance of $\sigma_0^2(\mathbf z_m)$ at $\mathbf z_m$ when we start the $k$-times' sampling at it. Formally, we have 
\[
\sigma_{n-K}^2(\mathbf{z}_m)\leq \sigma_0^2(\mathbf{z}_m),
\] 
and hence by (\ref{decomp}), we have
\begin{equation}\label{upbd}
\sigma_n^2(\mathbf{z}_m)\leq \frac{\sigma_{\epsilon}^2}{K+\sigma_{\epsilon}^2/\sigma_0^2(\mathbf z_m)}\leq \frac{\sigma_{\epsilon}^2}{K+\sigma_{\epsilon}^2/\overline{\sigma}^2}\quad \forall n\geq K,
\end{equation}
where $\overline{\sigma}^2:=\max_{i=1,\dots,m}\sigma_{0}^2(\mathbf z_i)$.

On the other hand, again by the monotonic decreasing property of the variance update (\ref{var_update}), we also see that $\sigma_{n-K}^2(\mathbf z_m)$ is no smaller than when all other grid points are sampled infinitely many times. Formally, we first notice that $\sigma_{n-K}^2(\mathbf z_m)$ is no smaller than whenever all other grid points are sampled $n-K$ times. Now applying exchangeability, associativity and monotonicity, we can take the limit that the number of samples collected at $\mathbf{z}_i$ goes to $\infty$ one by one from $i=1$ to $m-1$, and obtain that $\sigma_{n-K}^2(\mathbf z_m)$ is lower bounded by the (limiting) posterior variance quantity $s_{m-1}^{m,m}$ define/obtained as follows (denoting by $k_{n_1,\dots,n_m}$ the posterior covariance of sampling $\mathbf{z}_i$ for $n_i$ times as computed and defined by Eqns. (\ref{eqn:GP})):
\[
\begin{split}
s_0^{i,j}&:=k_{n-K,n-K,\dots,n-K,K}(\mathbf{z}_i,\mathbf{z}_j)\quad \forall i,j=1,\dots,m,\\
s_1^{i,j}&:=\lim_{n_1\rightarrow\infty}k_{n_1,n-K,\dots,n-K,K}(\mathbf{z}_i,\mathbf{z}_j)=s_0^{i,j}-\frac{s_0^{i,1}s_0^{1,j}}{s_0^{1,1}}\quad \forall i,j=2,\dots,m,  \dots, \\
s_l^{i,j}&:=\lim_{n_l\rightarrow\infty}k_{\infty,\dots,n_l\dots,n-K,K}(\mathbf{z}_i,\mathbf{z}_j)=s_{l-1}^{i,j}-\frac{s_{l-1}^{i,l}s_{l-1}^{l,j}}{s_{l-1}^{l,l}}\quad \forall i,j=l+1,m,  \dots,\\
s_{m-1}^{i,j}&:=\lim_{n_{m-1}\rightarrow\infty}k_{\infty,\dots,\infty,n_{m-1},K}(\mathbf{z}_i,\mathbf{z}_j)=s_{m-2}^{i,j}-\frac{s_{m-2}^{i,m-1}s_{m-2}^{m-1,j}}{s_{m-2}^{m-1,m-1}}\quad \forall i,j=m, 
\end{split}
\]
where we used the covariance update formula (\ref{post_formula}) and take the limit to remove the $\sigma_{\epsilon}^2$ term. Notice that although sampling infinitely many times is undefined in the probability sense, it is well-defined by Eqns. (\ref{eqn:GP}), or the update formulae \ref{post_formula} together with associativity and exchangeability. Readers should keep in mind that we are talking about the algorithmic convergence without any assumptions on the underlying probability models -- this is also one of the strengths of our algorithm, which is robust to model misspecification.

Here $s_l^{i,j}$ is exactly updating from the (limiting) posterior covariance $s_{l-1}^{i,j}$ by formula (\ref{post_formula}), with $n=\infty$ and $k_0(\mathbf{z}_i,\mathbf{z}_j)$ replaced by $s_{l-1}^{i,j}$, and $\bar{\mathbf{x}}$ replaced by $\mathbf{z}_l$.

Comparing with the well-known formula for conditional covariance of multivariate normal distributions, and utilizing the associativity of conditioning, we immediately see that $[s_l^{i,j}]_{i,j=l+1,\dots,m}$ is the posterior covariance matrix of the random vector $[f(\mathbf{z}_{l+1:m})]^T$ conditioned on $f(\mathbf{z}_{1:l})$ (without noises). In particular, we have 
\[
\sigma_{n-K}^2(\mathbf{z}_m)\geq s_{m-1}^{m,m}=\boldsymbol{\Sigma}^0_{m,m}-\boldsymbol{\Sigma}^0_{m,1:m-1}\left(\boldsymbol{\Sigma}^0_{1:m-1,1:m-1}\right)^{-1}\boldsymbol{\Sigma}^0_{1:m-1,m}:=\underline{\sigma}_m^2>0,
\]
where the last inequality comes from the fact that the Schur complement of a positive definite matrix is positive definite. In particular, since we have assumed that $k_0$ is a positive definite kernel, $\boldsymbol{\Sigma}^0\succ 0$ and hence the scalar Schur complement $\underline{\sigma}_m^2$ above is strictly positive.

Now recall that $\mathbf{z}_m$ is actually an arbitrary $\mathbf{z}_i$ (which we denote as $\mathbf{z}_m$ for notational convenience), defining $\underline{\sigma}:=\min_{i=1,\dots,m}\underline{\sigma}_m$, we immediately arrive at a lower bound on $\sigma^2_{n-K}(\mathbf{z}_m)$ by using associativity and formula (\ref{decomp}):
\begin{equation}\label{lwbd}
\sigma_{n}^2(\mathbf{z}_m)\geq \frac{\sigma_{\epsilon}^2}{K+\sigma_{\epsilon}^2/\underline{\sigma}^2} \quad \forall n\geq K.
\end{equation}

Combining (\ref{upbd}) and (\ref{lwbd}), we see that the posterior variance $\sigma_n^2(\mathbf{z}_m)$ after sampling $\mathbf z_m$ for $K$ times within $n(\geq K)$ timesteps is $\Omega(1/K)$ as $K\rightarrow\infty$ . Notice that this means that the order of magnitude of the posterior variance does not depend on $n$.

Finally, if we let Algorithm \ref{alg:Framwork} run forever with $\mathbf z_m$ sampled at most $T$ times then there must be at least one point in the grid, say $\mathbf z_1$, which is sampled infinitely often. For such a point the posterior variance $\sigma^2_{GP}(\mathbf z_1)$ tends to zero as the number of samples collected for $\mathbf z_1$ goes to $\infty$ by (\ref{upbd}) and (\ref{lwbd}), which implies that eventually the acquisition function $E_{GP}(\mathbf z_1) \leq u(\sigma_{GP}(\mathbf z_1))$ tends to zero by the hypothesis of the lemma. However, \[
E_{GP}(\mathbf z_m) \geq l\left(\sigma_{\epsilon}^2/(T+\sigma_{\epsilon}^2/\underline{\sigma}^2)\right)
\] is bounded away from zero because, by hypothesis of the lemma, $l(\cdot)$ is strictly increasing with limit $0$ at $0$, and we argued that $\underline{\sigma}^2$ is bounded away from zero. That is, eventually $E_{GP}(\mathbf z_m) > E_{GP}(\mathbf z_1)>0$ for all sufficiently large steps in which we collect at $\mathbf z_1$.  
This is a contradiction because we are assuming that at every step the next sample is chosen such that it maximizes the acquisition function, which implies that the choice of $\mathbf z_1$ is incorrect here.
 Hence, every grid-point must be sampled infinitely often. This completes our proof. 
\qed\\
}

\noindent\textbf{Lemma \ref{lemma:Lookahead_convergence}.}
For the acquisition function (\ref{eqn:Lookahead_Convergence}) with $\gamma > 0$, $\epsilon > 0$, we have:
\begin{itemize}
\item $u(\sigma_{GP}(\mathbf x^+)) = \max \left( |\Omega| \Phi \left(\frac{\sigma_{\epsilon}}{\overline{\sigma} \sigma_{GP}(\mathbf x^+)} \left(-\epsilon/2 \right) \right),\gamma \sigma_{GP}(\mathbf x^+) \right)$ 
\item $l(\sigma_{GP}(\mathbf x^+)) = \gamma \sigma_{GP}(\mathbf x^+)$ 
\end{itemize}
Also the lower bound $ l(\sigma_{GP}(\mathbf x^+))$ is monotonically increasing and 
$\lim_{\sigma(\mathbf x^+)\to 0^+} u(\sigma(\mathbf x^+)) = \lim_{\sigma(\mathbf x^+)\to 0^+} l(\sigma(\mathbf x^+)) = 0$.\\

\noindent\textbf{Proof:} 
Consider the acquisition function used in the paper:
\begin{eqnarray}
E_{GP}(\mathbf x^+) = \max(\E_{y^+} \left|I_{GP^+}\right| -  \left|I_{GP}^\epsilon \right| , \gamma \sigma_{GP}(\mathbf x^+) ) 
\end{eqnarray}

Clearly the lower bound $l(\sigma_{GP}(\mathbf x^+)) = \gamma \sigma_{GP}(\mathbf x^+)$ which is positive, strictly monotonically increasing in $\sigma_{GP}(\mathbf x^+)$, and goes to zero as $\sigma_{GP}(\mathbf x^+)$ goes to zero.
For the upper bound, assume $\E_{y^+} \left|I_{GP^+}\right| -  \left|I_{GP}^\epsilon \right| > \gamma \sigma_{GP}(\mathbf x^+)$ (otherwise $\gamma \sigma_{GP}(\mathbf x^+)$ is also the upper bound). 
We consider one term in the summation.
Suppose that  $\mu_{GP}(\mathbf x) - \beta\sigma_{GP}(\mathbf x) > t - \epsilon$, which is one of the two cases we will discuss. Then such a term can be expressed as:

\begin{equation}
\begin{split}
1-\Phi& \left(-\frac{\sqrt{\sigma^2_{GP}(\mathbf x^+) + \sigma^2_{\epsilon}}
 }{|\textup{Cov}_{GP}(f(\mathbf x), f(\mathbf x^+))|} \times \left( \mu_{GP}(\mathbf x) - \beta\sigma_{GP^+}(\mathbf x) -t  \right) \right)  \\
&-\mathbbm{1}\{\mu_{GP}(\mathbf x) - \beta\sigma_{GP}(\mathbf x) > t - \epsilon \} \\
=& -\Phi \left(-\frac{\sqrt{\sigma^2_{GP}(\mathbf x^+) + \sigma^2_{\epsilon}}
 }{|\textup{Cov}_{GP}(f(\mathbf x), f(\mathbf x^+))|} \times \left( \mu_{GP}(\mathbf x) - \beta\sigma_{GP^+}(\mathbf x) -t  \right) \right) < 0.
\end{split}
\end{equation}
Now assume $\mu_{GP}(\mathbf x) - \beta\sigma_{GP}(\mathbf x) \leq t - \epsilon$.
By Cauchy-Schwartz inequality, we have
$$|\textup{Cov}_{GP}(f(\mathbf x), f(\mathbf x^+))| 
\leq  \sigma_{GP}(\mathbf x^+) \sigma_{GP}(\mathbf x) \leq  \overline{\sigma} \sigma_{GP}(\mathbf x^+),$$ where the last passage follows from the fact that the variance is strictly decreasing and $\overline{\sigma}$ is an upper bound of the prior variances. By the update formulae in Eqns. (\ref{eqn:GP}), we have 
\begin{equation}
\begin{split}
\sigma_{GP}(\mathbf x) &- \sigma_{GP^+}(\mathbf x) \\
&= \sigma_{GP}(\mathbf x) \left( 1-\sqrt{1-\frac{\textup{Cov}^2_{GP}(f(\mathbf x), f(\mathbf x^+))}{(\sigma^2_{GP}(\mathbf x^+)+\sigma^2_{\epsilon})\sigma^2_{GP}(\mathbf x)}} \right)\\
&\leq \sigma_{GP}(\mathbf x) \frac{\textup{Cov}^2_{GP}(f(\mathbf x), f(\mathbf x^+))}{(\sigma^2_{GP}(\mathbf x^+)+\sigma^2_{\epsilon})\sigma^2_{GP}(\mathbf x)}\\
&\leq \frac{\sigma^2_{GP}(\mathbf x^+)\sigma_{GP}(\mathbf x)}{\sigma^2_{GP}(\mathbf x^+)+\sigma^2_{\epsilon}} \leq 
\frac{\sigma^2_{GP}(\mathbf x^+)\sigma_{GP}(\mathbf x)}{\sigma^2_{\epsilon}} \leq \frac{\overline{\sigma}\sigma^2_{GP}(\mathbf x^+)}{\sigma^2_{\epsilon}} \leq
\epsilon/(2\beta)
\end{split}
\end{equation}
for sufficiently small $\sigma^2_{GP}(\mathbf x^+)$. This implies that
$\mu_{GP}(\mathbf x) - \beta\sigma_{GP^+}(\mathbf x) \leq t - \epsilon$ can be recast as $\mu_{GP}(\mathbf x) - \beta\sigma_{GP^+}(\mathbf x) - t \leq - \epsilon + \beta\sigma_{GP}(\mathbf x) - \beta\sigma_{GP^+}(\mathbf x) < - \epsilon /2$.

Thus for sufficiently small $\sigma_{GP}(\mathbf x^+)$ the contribution of point $\mathbf x \in \Omega$ to the acquisition function is
\begin{equation}
\begin{split}
\Phi &\left(\frac{\sqrt{\sigma^2_{GP}(\mathbf x^+) + \sigma^2_{\epsilon}}
 }{|\textup{Cov}_{GP}(f(\mathbf x), f(\mathbf x^+))|} \times \left( \mu_{GP}(\mathbf x) - \beta\sigma_{GP^+}(\mathbf x) -t  \right) \right) \leq\\
&\leq \Phi \left(\frac{\sqrt{\sigma^2_{GP}(\mathbf x^+) + \sigma^2_{\epsilon}}
 }{|\textup{Cov}_{GP}(f(\mathbf x), f(\mathbf x^+))|} \times (-\epsilon / 2) \right)\\
&\leq \Phi \left(\frac{\sigma_{\epsilon}}{|\textup{Cov}_{GP}(f(\mathbf x), f(\mathbf x^+))|} \times (-\epsilon / 2) \right)\\
&\leq \Phi \left(\frac{\sigma_{\epsilon}}{\overline{\sigma} \sigma_{GP}(\mathbf x^+)} \times (-\epsilon / 2) \right),
\end{split}
\end{equation}
where we use the monotonicity of the normal CDF.

Thus an asymptotic upper bound for $\E_{y^+} \left|I_{GP^+}\right| -  \left|I_{GP}^\epsilon \right| $  that depends only on the variance of the current sampling location is $ |\Omega| \Phi \left(\frac{\sigma_{\epsilon}}{\overline{\sigma} \sigma_{GP}(\mathbf x^+)} \left(-\epsilon/2 \right) \right)$.
By continuity, we have
$$
\lim_{ \sigma_{GP}(\mathbf x^+) \rightarrow 0^+} |\Omega| \Phi \left(\frac{\sigma_{\epsilon}}{\overline{\sigma} \sigma_{GP}(\mathbf x^+)} \left(-\epsilon/2 \right) \right)\rightarrow 0
$$
which implies that $E_{GP}(\mathbf x^+) \rightarrow 0$ as $\sigma_{GP}(\mathbf x^+) \rightarrow 0$
\qed\\

\section*{Appendix B: Safe Exploration}
To the best of our knowledge the available algorithms for level set estimation assume that $f$ can be evaluated at an arbitrary point. However, this is not always the case. In medical applications, for example, experiments need to be done on humans and therefore it is a requirement to never query the function below the threshold, which can result in death of the patient. One needs to act conservatively and would not explore the domain far away from known regions, unless prior knowledge is available. The GP that models the function can be used to identify regions which are not deemed safe, that is, we can restrict the sampling location to the set of safe points under the current GP, which are the points in the set $S$:
\begin{equation}
S = \{ \mathbf x \mid \mu_{GP}(\mathbf x) - \gamma\sqrt{\sigma_{GP}^2+\sigma_{\epsilon}^2} > t \}.
\label{safethresh}
\end{equation}  
In Eqn. (\ref{safethresh}), we include the noise $\epsilon \sim \mathcal{N}(0,\sigma^2_{\epsilon})$ as we do not want the outcome $y^+$ to be below the threshold. If one is not concerned with the noise and sets $\gamma = \beta$ then the set of candidate points coincides with $I_{GP}$. Algorithm \ref{alg:ConceptualLookahead} can then be changed in a straightforward manner by only considering points in the ``safe region" $S \subset \Omega$ as potential query points.

\begin{figure}[h]
\centering
\includegraphics[width=3.8cm]{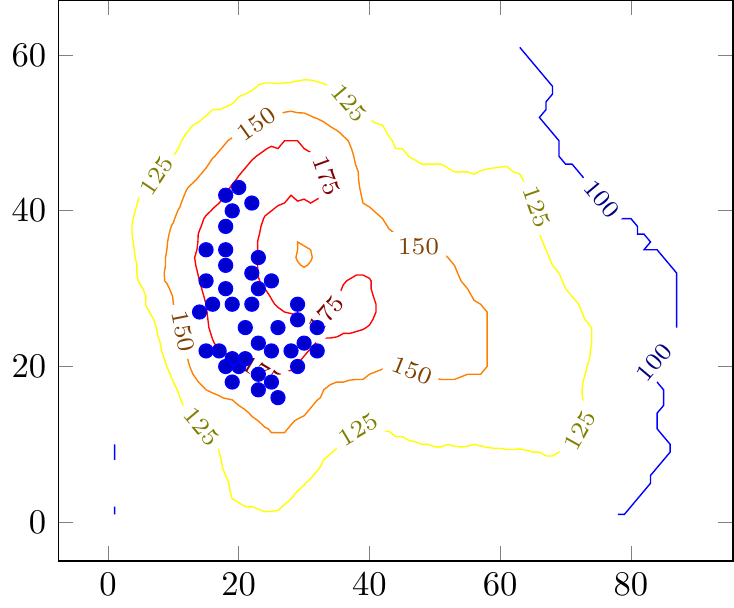}
\includegraphics[width=4cm]{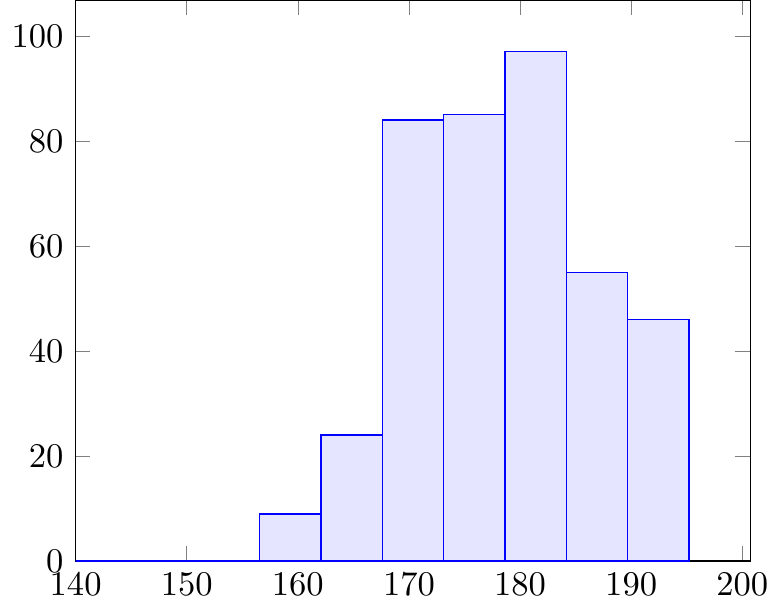}
\caption{Left picture: sampling locations at an intermediate step during the execution of the algorithm. Notice that no sample is too close to the threshold $t = 150$. Right picture: distribution of the outcomes at the points selected by the algorithm during 10 runs with a budget of 40 points.}
\label{fig:volcanosafe}
\end{figure}

We examine the topographic dataset of the Auckland's Maunga Whau Volcano, which is available on a 10m by 10m grid on the R \texttt{datasets}\footnote{\url{http://stat.ethz.ch/R-manual/R-devel/library/datasets/html/00Index.html}} package. 
We run 10 simulations in the available $87 \times 61$ measurements, with noises added to the evaluations ($\log(\sigma_{\epsilon}) = -1.0$).
We fine-tune the GP hyper-parameters $\sigma_{ker}, l$ on a small number of held-out samples via maximum likelihood estimation and use them in the GP prior. We want to identify the threshold $t = 150$ and give $\mathbf x = [20, 20]$ as the initial seed for the algorithm. Notice that it is necessary to provide such a starting point above the threshold or an appropriate prior GP that results in at least one point in the set of ``safe'' points $S \subset \Omega$ so that the algorithm can explore. Figure 
\ref{fig:volcanosafe} reports the location of $40$ points sampled by the algorithm in one run. For simplicity we set $\gamma = \beta = 1.96$. Notice that there are no points below $t = 150$, as one would expect. We further report in the same figure the distribution of the sampled values over the 10 runs; the algorithm reaches out to the boundary around $160$, but hardly ever samples points near the threshold of $150$.

\end{document}